\documentclass[preprint, times, sort&compress]{elsarticle}
\journal{Journal of \LaTeX\ Templates}

\usepackage{amsthm}
\usepackage{graphicx}
\usepackage{amsmath}
\usepackage{mathtools}
\usepackage{enumitem}

\usepackage[ruled,linesnumbered,resetcount]{algorithm2e}
\usepackage{algpseudocode}
\usepackage{booktabs}
\usepackage[table]{xcolor}
\usepackage{pdflscape}
\usepackage{longtable}
\usepackage{multirow}
\usepackage{geometry}
\usepackage{threeparttablex}
\usepackage{graphicx}
\usepackage[compatibility=false]{caption}
\usepackage[labelformat=simple]{subcaption}
\usepackage[numbers]{natbib}
\usepackage{rotating}

\usepackage{lipsum}
\usepackage{import}
\usepackage[acronym]{glossaries}
\makeglossaries

\usepackage{graphicx}
\usepackage{amsmath}
\usepackage{mathtools}
\usepackage[ruled,linesnumbered,resetcount]{algorithm2e}
\usepackage{algpseudocode}

\usepackage{pdfpages}
\usepackage{import}
\usepackage{multicol}
\usepackage{tabularx}
\usepackage{rotating}
\usepackage{float}

\biboptions{longnamesfirst, compress}
\newacronym{ga}{GA}{Genetic Algorithm}
\newacronym{ea}{EA}{Evolutionary Algorithm}
\newacronym{mfo}{MFO}{Multifactorial Optimization}
\newacronym{mfea}{MFEA}{Multifactorial Evolutionary Algorithm}
\newacronym{cluspt}{CluSPT}{Clustered Shortest-Path Tree Problem}
\newacronym{sto}{STO}{single-tasking of Genetic Algorithm}
\newacronym{clumrct}{CluMRCT}{Minimum Routing Cost Clustered Tree Problem}
\newacronym{clutsp}{CluTSP}{Clustered Traveling Salesman Problem}
\newacronym{ftsp}{FTSP}{Family Traveling Salesman Problem}
\newacronym{gtsp}{GTSP}{Generalized Traveling Salesman Problem}
\newacronym{gp}{GP}{Genetic Programming}
\newacronym{ant}{ACO}{Ant Colony Optimization}
\newacronym{grasp}{GRASP}{Greedy Randomized Adaptive Search Procedure}
\newacronym{interclumrct}{InterCluMRCT}{Minimum Inter-cluster Routing Cost Clustered Tree Problem}

\newacronym{tsp}{TSP}{Traveling Salesman Problem}
\newacronym{steinertp}{STP}{Steiner Tree Problem}
\newacronym{clusteiner}{CluSteiner}{Clustered Steiner Tree Problem}
\newacronym{uss}{USS}{Unified Search Space}
\newacronym{de}{DE}{Differential Evolution Algorithm}
\newacronym{pso}{PSO}{Particles Warm Optimization Algorithm}
\newacronym{rmp}{rmp}{Probability of Random Mating}
\newacronym{vrp}{VRP}{Vehicle Routing Problem}
\newacronym{cvrp}{CVRP}{Capacitated Vehicle Routing Problem}
\newacronym{ltga}{LTGA}{Linkage Tree Genetic Algorithm}
\newacronym{mfed}{MFEA}{Multifactorial evolutionary algorithm}
\newacronym{rga}{RGA}{Randomized Greedy Algorithm}

\newacronym{mfltga}{MF-LTGA}{Multifactorial Linkage Tree Genetic Algorithm}

\newacronym{spta}{SPTA}{Shortest Path Tree Algorithm}
\newacronym{scm}{SCM}{Swap-Change Mutation operator}

\newacronym{clutp}{CluTP}{Clustered Tree Design Problem}
\newacronym{blop}{BLOP}{Bi-level Optimization Problem}
\newacronym{n-lsea}{N-LSEA}{Nested Local Search Evolutionary Algorithm}
\newacronym{m-lsea}{M-LSEA}{Multi-tasking Local Search Evolutionary Algorithm}

\newacronym{cesa}{CESA}{}
\newacronym{iim}{IIM}{New Individual Initialization Method}
\newacronym{ncx}{NCX}{New Crossover Operator}
\newacronym{nmo}{NMO}{New mutation Operator}

\newacronym{rpd}{RPD}{Relative Percentage Differences}
\newacronym{nea}{NEA}{New Evolutionary Algorithm}
\newacronym{s-cluspt}{S-CluSPT}{}

\newacronym{hbrga}{HBRGA}{Heuristic Based on Randomized Greedy Algorithm}
\newacronym{g-mfea}{G-MFEA}{New Multifactorial Evolutionary Algorithm}

\newcommand{\scalefigure}{0.20}

\newtheorem{definition}{Definition}
\newtheorem{proposition}{Proposition}

\begin{document}

\begin{frontmatter}
	\title{Evolutionary Algorithm and Multifactorial Evolutionary Algorithm on Clustered Shortest-Path Tree problem}
	
	\author[pthh]{Phan Thi Hong Hanh}
	\ead{mliafol.phan86@gmail.com}
	
	\author[pdt]{Pham Dinh Thanh}
	\ead{thanhpd05@gmail.com}
	
	\author[httb]{Huynh Thi Thanh Binh\corref{cor1}}
	\ead{binhht@soict.hust.edu.vn }
	
	\cortext[cor1]{Corresponding author.}
	
	\address[pthh]{Quantitative Hedge Fund, Singapore}
	\address[pdt]{Faculty of Natural Science - Technology, Taybac University, Vietnam}
	\address[httb]{School of Information and Communication Technology, Hanoi University of Science and Technology, Vietnam}
	
	\begin{abstract}
		In literature, \gls{cluspt} is an NP-hard problem. Previous studies focus on approximation algorithms which search for an optimal solution in relatively large space. Thus, these algorithms consume a large amount of computational resources while the quality of obtained results is lower than expected. In order to enhance the performance of the search process, this paper proposes two different approaches which are inspired by two perspectives of analyzing the \gls{cluspt}. The first approach intuition is to narrow down the search space by reducing the original graph into a multi-graph with fewer nodes while maintaining the ability to find the optimal solution. The problem is then solved by a proposed evolutionary algorithm. This approach performs well on those datasets having small number of edges between clusters. However, the increase in the size of the datasets would cause the excessive redundant edges in multi-graph that pressurize searching for potential solutions. The second approach overcomes this limitation by breaking down the multi-graph into a set of simple graphs. Every graph in this set is corresponding to a mutually exclusive search space. From this point of view, the problem could be modeled into a bi-level optimization problem in which the search space includes two nested search spaces. Accordingly, the \gls{n-lsea} is introduced to search for the optimal solution of gls{cluspt}, the upper level uses a simple Local Search algorithm while the lower level uses the Genetic Algorithm. Due to the neighboring characteristics of the local search step in the upper level, the lower level reduced graphs share the common traits among each others. Thus, the \gls{m-lsea} is proposed to take advantages of these underlying commonalities by exploiting the implicit transfer across similar tasks of multi-tasking schemes. The improvement in experimental results over \gls{n-lsea} via this multi-tasking scheme inspires the future works to apply gls{m-lsea} in graph-based problems, especially for those could be modeled into bi-level optimization.
	\end{abstract}
	
	
\end{frontmatter}

\glsresetall

\section{Introduction}
\label{sectisec_intro}
In science and engineering, the network design and optimization problems have attracted much interest from the research community. In several network infrastructure, a group of existing nodes might be considered as clusters, the objective is to optimize the connections among clusters while satisfying connecting characteristic between two arbitrary nodes in each cluster. This problem is called \gls{clutp}. One of the most well-known problems of \gls{clutp} is \gls{cluspt} which plays an important role in network design, products distribution, agricultural irrigation. With respects to network design, the \gls{cluspt} can be used to tackle the optimization problems in network topology, urban network, local area network interconnection, cable TV system and fiber. In agricultural irrigation system, \gls{cluspt} can be applied in optimizing irrigation system for the desert places where a source of water is shared with several areas. Each area is considered as a cluster, and the objective is to build an optimal system to bring water from the source to all important parts of each area. Similar to the irrigation system, \gls{cluspt} can be employed to build products distribution system which transmits goods from the top-level to the middle agencies, before, those goods are distributed to all stores. \gls{cluspt} helps the system to save cost and time.

Due to \gls{cluspt} being an NP-Hard problem~\cite{wu2015clustered, thanh_rga_2019}, exact algorithms are only suitable for small instances. To cope with larger instances, approximation approaches are more appropriate. \glspl{ea} have emerged and have been applied to solve many problems on both theoretical and practical research among various fields. In \glspl{ea}, a solution of the problem is considered as an individual, through natural selection, the individuals in the next generation tend to outperform the individuals in previous generations. Therefore, the solutions produced by \glspl{ea} will always be near or tend towards the optimal solutions. In this work, we use \gls{ea} as a foundation to further develop algorithms to deal with \gls{cluspt}.

Although \gls{ea} is really powerful in combinatorial optimization problems, the algorithm still consumes large amount of computing power in case the traversal space is huge. Therefore, in this paper, we analyze and solve the \gls{cluspt} in two different approaches of which objectives are to scale down the search domain. In the first approach, all network nodes of a cluster are contracted into a single vertex, then in the transformed graph, we apply \gls{ea} to evolve cluster interconnections. For each cluster, we employ an exact algorithm to obtain the local shortest path tree. In the second approach, we transform the search space of \gls{cluspt} into two nested search spaces corresponding to different optimization levels. In the upper level, we consider the combination of representative nodes for all clusters that each node is selected from each distinctive cluster. Depending on the chosen node for each cluster, a shortest path tree would be constructed by an exact algorithm that the node take place as the source vertex in the correlative cluster. In order to optimize the representative combination, the authors propose a search strategy based on the idea of hill-climbing algorithm. According to the selected nodes in the upper level, in the lower level, the goal is to optimize the spanning tree for the graph whose the edge set includes connections from clusters to the representative nodes and the vertex set is the representative combination. Several spanning tree optimization problems are generated in accordance with corresponding representative nodes, these individual problems are optimized through \gls{ea}. Hence, \gls{n-lsea}, which amalgamates the hill-climbing and evolving phases, is proposed to fulfill this approach.

When diving deeper into solving several spanning tree optimization problems, we relize that these sub-problems have high similarity which is indicated by the high probability of common edges among the achieved individual spanning trees. This investigation convinces us to apply the idea of \gls{mfo} to take the advantage of meaningful building-blocks shared between different optimization tasks \cite{gupta2016multifactorial}. \gls{mfo}, which is an optimization paradigm involving many optimization tasks all together, was introduced and applied to a numerous optimization problems covering combinatorial and continuous problems \cite{ong2016evolutionary,zhou2016evolutionary,yuan2016evolutionary,chandra2016evolutionary}. In terms of dealing with \gls{mfo}, \gls{mfea}~\cite{trung2019multifactorial, binh2020multifactorial} was proposed as a more general form of \gls{ea} \cite{ong2016evolutionary}, which added more factors for assessing selection pressure. By considering several spanning tree optimization problems as a \gls{mfo} problem, we introduce \gls{m-lsea}, which is an advanced version of \gls{mfea} applying local refinement methods.

\indent Accordingly, the main contributions of this research are:
\begin{itemize}
	\item Propose proper \gls{ea} based algorithm using the first approach that the performance is better than previous heuristic methods in medium size data instances.
	\item Propose \gls{ea} based improvement \gls{n-lsea} by speculating the \gls{cluspt} as a combinatorial optimization as mentioned in the second approach.
	\item Propose \gls{mfea} based algorithm \gls{m-lsea} which further excel the performance in larger data instances because exploiting the similarities in solving multiple spanning tree optimization problems concurrently.
\end{itemize}
The performance for mentioned algorithms are evaluated on a generated dataset. Due to the fact that there are no available test suites for \gls{cluspt}, in this paper, we generate the test sets based on the MOM\_lib \cite{helsgaun_solving_2011, mestria_grasp_2013} of the Clustered Traveling Salesman Problem.

The rest of this paper is organized as follows. In Section \ref{sec_definition}, the detail definition of the \gls{cluspt} is formulated and other notations and definitions are also provided. Section \ref{sec_related} discusses about related works. The proposed genetic algorithm for the \gls{cluspt} is elaborated in Section \ref{sec_proposal}. Section \ref{sec_metric} proposes an exact method for solving the \gls{cluspt} on metric graphs. Section \ref{sec_model} explains the way to transform \gls{cluspt} search space to two nested search spaces and our proposal algorithms for generated search spaces. Section \ref{sec_results} explains the setup of our experiments and reports the computed results. The paper concludes in section \ref{Sec_Conclusion} with discussions on the future extension of this research.

\section{Problem formulation}
\label{sec_definition}
In this paper, for a graph $G(V, E)$ $V$ and $E$ denote the vertex set and edge set of $G$, respectively. \\
Let $G = (V,E,w)$ be an undirected weighted graph. The function weight is w which each edge $(u,v) \in E$ has a non-negative weight $w(u,v)$. An arbitrary path from a vertex $u$ to vertex $v$ on $G$ is simple if every edge on this path appears one time. In this paper, we only consider simple path. Clearly, on a tree, between an arbitrary pair vertex $u$, $v \in V(T)$, there is an unique simple path and the length of this path is marked as $d(u,v)$. Consider a vertex subset $S \subseteq V$, $G[S]$ is used to stand for the sub-graph of $G$ induced by $S$. For a cluster problem, the set vertex $V$ is partitioned into $k$ clusters, symbolized by $C_1,C_2,\ldots,C_k$ satisfied $\bigcup_{i= 1}^{k}{C_i} = V $ and $C_i \cap C_j = \emptyset$ $(i \neq j)$. A spanning tree $T$ on $G$ is marked as a clustered spanning tree of G if and only if, with each cluster $i$, $T$ is a tree.

The definition of \gls{cluspt}~\cite{huynh2020multifactorial, d2019hardness} is given as followed:

\noindent\textbf{Input}: Given an undirected weighted graph $G = (V,E,w)$ whose vertex set is divided into $k$ disjointed clusters $C=\{C_1,C_2,\dots, C_k\}$.  $G$ has a root, denoted by $r$.\\
\textbf{Output}: A clustered spanning tree $T$ of graph $G$. \\
\textbf{Objective}: Minimize $Cost=\sum_{v \in V}{d(r,v)}$.

\begin{definition}
	Consider a tree $T$ and a vertex subset $S \subseteq V(T)$, the local tree of $S$ on $T$ is $T(S)$. For simplicity, in this paper, local trees of a clustered spanning tree $T$ are trees which reduced by all clusters. A rooted tree is a tree in which one vertex of this tree is designated as the root. Consider two nodes $u$, $v$ in a rooted tree $T$ and the root is $r$:
	\begin{itemize}
	    \item Node $u$ is called the parent of $v$ or $v$  is child of $u$, if and only if $(u,v)$ is an edge in the path $r$ to $v$. If $u$ is the parent of $v$ then $d(r,v) = d(r,u) + w(u,v)$.
	    \item Node $u$ is called the ancestor of $v$ or $v$ is descendant of $u$, if and only if the path from $r$ to $v$ passes through $u$. We use the notation $N(u)$ to denote the set of all vertices that are descendants of $u$ on tree $T$.
	\end{itemize}
\end{definition}

\begin{definition}
	Let $G = (V, E, w)$ be a 
	graph, $C = \left\{C_1, C_2,\ldots, C_k\right\}$ be a partition of $V$ and $T$ be a clustered spanning tree of $G$, $T$ is a rooted tree whose root is $r$. Consider an arbitrary cluster $C_t (1 \leq t \leq k)$, a vertex $r_t \in V$ is called the root of the local tree of $C_t$ on $T$ if $r_t \in C_t$ and $\forall u \in C_t$, $u \neq r_t$, $u$ is the descendant of $r_t$ on $T$. \\
\end{definition}

\begin{definition}
	Let $G = (V, E, w)$ be a graph, $C = \left\{C_1, C_2,\ldots, C_k\right\}$ be a partition of $V$ and $T$ be a cluster spanning tree of $G$, $T$ is a rooted tree whose root is $r$. Consider two vertices $u, v \in V$ ($u \in C_i, v \in C_j, 1 \leq i, j \leq k$), the edge  is designated as an inter-cluster edge if and only if $u, v$ are in distinguish cluster ($i \neq j$).  Clearly, if $u$ is the parent of $v$, then $v$ is the root of the local tree of $C_j$ and $u$ is denoted as one port of $C_j$. \\  
	Let $G = (V, E, w)$ be a graph, $C = \left\{C_1, C_2,\ldots, C_k\right\}$ be a partition of $V$, $r$ be the root of $G$. With no loss of generality, we assume that $r \in C_1$. \\
	Let $T$ be a cluster spanning tree of $G$, $T$ is a rooted tree whose root is $r$. $T_1, T_2,\ldots, T_k$ are the local trees of $C_1, C_2,\ldots, C_k$ on $T$ and $r_1, r_2,\ldots, r_k$ are roots of $T_1, T_2,\ldots, T_k$, respectively.
\end{definition}

\begin{figure}[htbp]
	\centerline{\includegraphics[scale=0.3]{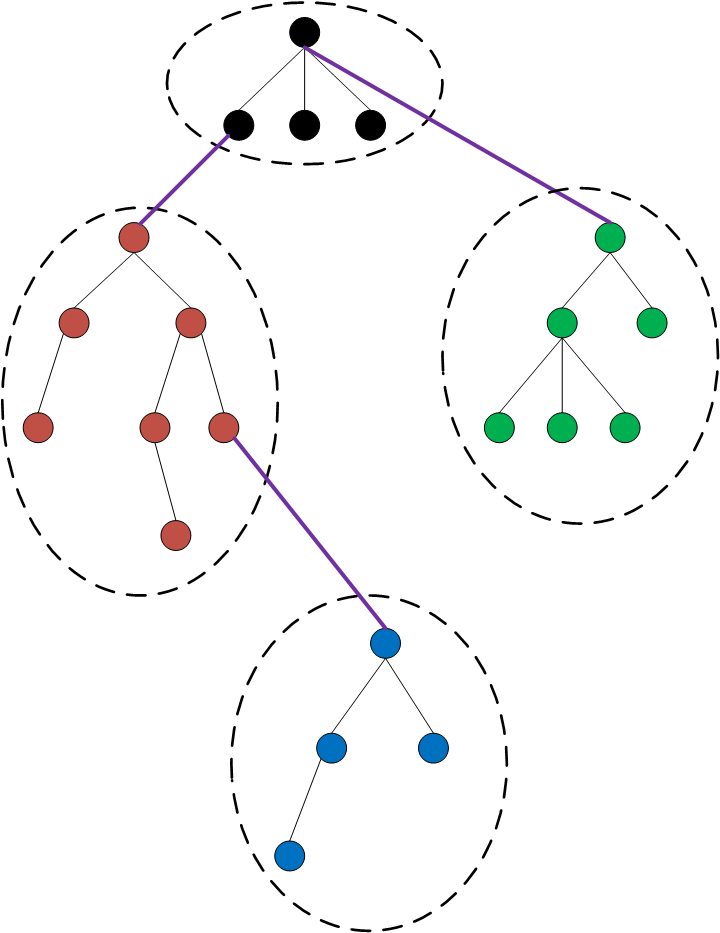}}
	\caption{A sample of the cluster spanning tree. There are four local trees in circles, one for each cluster. $r_1, r_2, r_3, r_4$ are the roots of those local trees. There are three inter-cluster edges, symbolized as purple lines. $(u, r_2)$ is an inter-cluster edge, and $u$ is a port of cluster $C_1$.}
	\label{fig_01}
\end{figure}

\begin{proposition}
	\label{pro:1}
	If $T$ is an optimal solution for graph $G$, then $T_1, T_2,\ldots, T_k$ are the shortest path tree rooted at $r_1, r_2,\ldots, r_k$~\cite{wu2015clustered} of $G$. A shortest path tree rooted at $v$ is a spanning tree which distance from $v$ to any node $u$ in tree is the shortest path distance from $v$ to $u$ in $G$.
\end{proposition}

\section{Related works}
\label{sec_related}
Many applications in real-life relate to network optimization problems \cite{myung1995generalized}. Among network problems there are many variants of which network structures are clusters \cite{myung1995generalized, dror2000generalized, prisco1986fiber}.

Bang~Ye~Wu~et~al.~\cite{wu2015clustered} considered a version of the Steiner Minimum Tree, \gls{clusteiner} in which vertices are partitioned into clusters. To solve \gls{clusteiner}, an approximation algorithm is proposed. In the new algorithm, Hamiltonian paths are added to all local tree of clusters, then edges which belong to inter-cluster topologies are removed. Bang~Ye~Wu~et~al. also pointed out that Steiner ratios can get maximum at 4 and minimum at 3 and the new algorithm can be approximated in $(2 + \rho)$ for \gls{clusteiner} in $O(nlogn + f(n))$ time, when local topologies are given, \gls{clusteiner} can be $(1 + \rho)$-approximated in $O(nlogn + f(n))$ time. 

Chen-Wan~Lin and Bang~Ye~Wu~\cite{lin2017minimum} studied a new variant of clustered tree, Minimum Routing Cost Clustered Tree Problem~(CluMRCT). A new constraint in CluMRCT is that sub-graph in each clustered is spanning tree. The authors demonstrated that a CluMRCT has more than 2 clusters is NP-hard and proposed an approximation algorithm for solving the CluMRCT. In the new algorithm, the solution of CluMRCT is found by creating a two level star-like graph and based on the R-star graph. The authors demonstrated that the routing cost of R-star graph is less than twice times the cost of the optimal solution of CluMRCT problem. In~\cite{myung1995generalized}, the authors studied a generalized version of the Minimum Spanning Tree Problem (GMSTP). The vertices were partitioned into distincted groups. The goal was to find a minimum-cost tree which contains only one vertex from each group. After proving that GMSTP was NP-hard, the authors proposed two mathematical programming formulations and compared them in terms of linear programming relaxations when applied to GMSTP.

D’Emidio~et~al.~\cite{demidio_clustered_2016} studied \gls{cluspt}, another version of clustered problems. The \gls{cluspt} can be found in many real-life network optimization problems such as network design, cable TV system and fiber-optic communication. The authors proved that the \gls{cluspt} was NP-Hard and proposed an approximation algorithm (hereinafter AAL) for solving the \gls{cluspt}. The main idea of the AAL is find a minimum spanning tree of each cluster and create a new graph by considering each cluster as a vertex. 

Recently, some types of algorithms are proposed to deal with the \gls{cluspt} i.e. greedy algorithm, \gls{ea} and \gls{mfea}. two new \gls{mfea} algorithms are used for solving the \gls{cluspt}. In~\cite{ThanhPD_TrungTB}, the authors proposed an \gls{mfea} (E-MFEA) with a new evolutionary operator for finding the solution of \gls{cluspt}. The major idea of these evolutionary operators is that first constructing spanning tree for the smallest sub-graph then the spanning tree for larger sub-graph are construed which based on the spanning tree for smaller sub-graph. In~\cite{ThanhPD_DungDA}, the authors took the advantage of Cayley code~\cite{thompson2007dandelion, perfecto2016dandelion, julstrom2005blob, palmer_representing_1994, paulden_recent_2006} to encode the solution of \gls{cluspt} and proposed evolutionary operators based on Cayley code. The genetic operators are constructed based on ideas of operators for binary and permutation representation.
In~\cite{binh_new_2019}, a \gls{nea} using a combination between \gls{ea} and Dijsktra's algorithm is proposed. In this approach, the \gls{cluspt} is decomposed into two sub-problems: the first sub-problem is to determine a spanning tree which connects the clusters together, while the second sub-problem is to determine the best spanning tree for each cluster. The main objective of \gls{ea} is to find the solution of the first sub-problem while the main objective of Dijkstra's Algorithm is to determine the solution of the second problems. Although \gls{nea} reduces the consumed resource and time for finding the \gls{cluspt} solution, some disadvantages also exist in \gls{nea}, i.e. a cluster connects to other clusters through only one vertex, so in some cases the obtained solution is not optimized. In~\cite{thanh_rga_2019}, the authors introduced a \gls{hbrga} to deal with the \gls{cluspt}. The idea of \gls{hbrga} is a combination of \glspl{rga} and Dijkstra's Algorithm. In \gls{hbrga}, the shortest-path tree for each cluster is constructed by Dijkstra's Algorithms while the edges connecting the clusters are built by \glspl{rga}. The major advantage of \gls{hbrga} is that it capable of exploiting search space to produce better solutions starting from the initial solution. However, \gls{hbrga} is prone to be trapped in local optimums when the problem's dimension increases since greedy-based algorithms have low exploration capability. In~\cite{thanh2020efficient}, authors described a \gls{g-mfea} having two tasks to deal with the \gls{cluspt} in which \gls{cluspt} solutions is determined in the first task while the objective of the second task is to improve quality of a part of \gls{cluspt} solutions in the first task. The main distinction between \gls{g-mfea} and other previous studies of \gls{mfea} is that, in those studies, each task will be finding the solution of a different problem. While \gls{g-mfea} solves a problem by dividing it into two tasks where the first one focus on solving the original problem and the other one solves a problem decomposed from the original problem. The paper shows that, \gls{g-mfea} outperformed existing heuristic algorithms in most of the test cases.

Many problems in real-life have special structure where decision-making of that can be made in hierarchical order. A problem that has two levels (higher and lower level decision-makers) in a hierarchy is called bi-level optimization problem (BLP). In other words, BLP is a nested optimization problem. BLP has many applications in theoretical research and real-life applications such as traffic planning~\cite{migdalas1995bilevel}, eco-traffic signal system~\cite{jung2016bi}, HVAC system~\cite{zhuang2017bi}, transportation network design, design of local area networks~\cite{camacho2015genetic, bard2013practical}, production planning~\cite{bard2013practical}, etc.

Bi-level optimization is a special instance of optimization in which a problem is embedded (nested) within another. An outer optimization task is commonly referred to as an upper-level optimization task, and an inner optimization task is commonly referred to as a lower-level optimization task. These problems involve two types of variables, referred to as the upper-level variables and the lower-level variables.

Although the BLP has confused structure, various algorithms are proposed to solve the BLP which include: exact algorithms (branch-and-bound~\cite{hansen1992new}, etc.), approximation algorithms (genetic algorithm~\cite{camacho2015genetic, oduguwa2002bi}, particle swarm optimization~\cite{zhang2012improved, gao2008particle}, differential evolution \cite{koh2007solving, angelo2013differential, angelo2014differential}, memetic algorithm~\cite{islam2015memetic}, simulated annealing algorithm~\cite{zhang_improved_2016}, etc.). \gls{ga} is one of the most effective algorithms for solving the BLP.

\gls{ga}, a computational model inspired by evolution, has been applied to a large number of real world problems~\cite{agoston_eiben_2003}. \gls{ga} can be used to get global solutions of problems~\cite{agoston_eiben_2003}. High adaptability and the generalizing feature of \gls{ga} help solve many problems by a simple structure. 

One of the first evolutionary algorithms for solving BLPs was proposed in the early 1990s. Mathieu~et~al.~\cite{mathieu1994genetic} proposed a new algorithm (GABBA) based on \gls{ga} and linear programming for solving BLP. GABBA only uses mutation operator and \gls{ga} is used in leader’s decision vector with individual representation based on string of base-10 digits. Another approach was proposed in~\cite{yin2000genetic} where \gls{ga} was used in upper level and  Frank-Wolfe algorithm was used in lower level. The main advances of the proposal algorithm (GAB) are that GAB can find the global optimum and the structure of GAB is very simple for implementation. 

In recent years, new algorithms based on \gls{ea} has been being proposed such as: memetic algorithm~\cite{islam2015memetic}, Linkage Tree Genetic Algorithm~\cite{thierens2013hierarchical}, Evolutionary multitasking algorithm~\cite{gupta2016multifactorial,gupta2015evolutionary,bali_multifactorial_nodate}, etc., in which, the evolutionary multitasking algorithm has emerged as a powerful algorithm to solve various complex problems in the research fields. Evolutionary multitasking algorithm was proposed by Abhishek Gupta, et al.~\cite{gupta2016multifactorial}. The idea of \gls{mfo} is borrowed from bio-cultural models of multifactorial inheritance. A difference between \gls{mfo} and \gls{ga} is that \gls{ga} can only solve a problem while \gls{mfo} can solve multiple problems at a time. To solve multiple problems at a time, \gls{mfo} uses a commons representation in unified search space for all problems and a solution of a specific problem can be achieved by transforming a solution in the unified search space. The performance of \gls{mfo} is verified on both discrete and continuous problems. The obtained results pointed out that \gls{mfo} has many advantageous features compared with single task.

Feng,~L~et~al.~\cite{feng2017empirical} proposed multifactorial optimization for particle swarm optimization (called multifactorial particle swarm optimization - MFPSO) and differential evolution search (called multifactorial differential evolution - MFDE). Due to particle swarm optimization (PSO) and differential evolution (DE) use only an individual for reproduction or update operations, this paper focuses on finding the assortative mating schemes for the new algorithms.  In MFPSO, a particle $i$ using a global best solution which has different skill factors from particle $i$ for updating its velocity on rmp rate. In MFDE, a trial vector $v_q^i$ is computed based on two random individuals having different skill factors from an individual $x_q^i$ on rmp rate. Mei-Ying~Cheng,~et~al.~\cite{cheng2017coevolutionary} proposed \gls{mfo} (named MT-CPSO) based on cooperative co-evolution framework. In MT-CPSO, particle swarm algorithm is used as a sample instantiation of a base optimizer for a real-parameter unification scheme. Two main differences between MT-CPSO and the existing \gls{mfo} are the knowledge transfer mechanism and interval for exchanging of knowledge. Distincted tasks in MT-CPSO are tackled by distincted subpopulations and  if the best results of a particle of a subpopulation is unchange in prescribe number of interactions then useful information from other subpopulations is across to help finding more promising regions in the unified search space. In~\cite{gupta2015evolutionary}, Gupta,~A.~et~al. proposed \gls{mfea} in basic Nested Bi-Level Evolutionary Algorithm (N-BLEA) by using the \gls{mfea}~\cite{dinh2020multifactorial} in lower level. A difference between novel algorithm (called M-BLEA) and  N-BLEA is that standard \gls{ea} is replaced by \gls{mfo} in performing lower level. In M-BLEA, the nearest individuals in each generation of upper level \gls{ea} are partitioned into groups based on a metric which determines relationship between individuals. After that M-BLEA performs \gls{mfo} for each group in lower level \gls{ea}.

\section{\glsentryshort{cluspt} on metric graphs}
\label{sec_metric}

\section{Proposed evolutionary algorithm}
\label{sec_proposal}
According to proposition~\ref{pro:1}, if there is a combination of all inter-cluster edges, we can find the rest of the edges to construct a solution for \gls{cluspt} such that its objective is optimized. To find a combination of inter-cluster edges, we first create a multi graph $G' = (V', E')$ based on the graph $G$ which:
\begin{itemize}
    \item[$-$] Each vertex in $V'$ stands for a cluster in $G$. $V' = \left\{v_1, v_2,\ldots, v_k\right\}$ and $v_i$ stands for $C_i(1 \leq i \leq k)$.
    \item[$-$] If there is one edge $e = (u, v, c) \in E$ which $u \in V_i, v \in V_j \ (i \neq j)$, then $v_i$ and $v_j$ are connected by an edge $e = (u, v, c)$.
\end{itemize}
$G'$ is called the clustered graph of $G$. Denote $D$ as the set of all spanning trees of $G'$. The set $D$ includes all feasible combinations of all inter-cluster edges. 

Consider a spanning tree $u \in D$, we  can construct an optimal cluster spanning tree $T_u$ of $G$ base on $u$ by following steps:
\begin{itemize}
    \item[$-$] All edges of $E(u)$ are selected to be inter-cluster edges in $T_u$.
    \item[$-$] By using Depth First Search algorithm, we can specify all roots of local trees of $T_u$, denoted as $r_{u1}, r_{u2},\ldots, r_{uk}$.
    \item[$-$] For each $i (1 \leq i \leq k)$, find the shortest path tree rooted at $r_{ui}$ of $G[C_i]$. Add all edges of this tree to $E(T_u)$.
\end{itemize}
The following example in Figure~\ref{fig:fig_02} illustrates these above steps.


\setlength{\intextsep}{0pt}
\renewcommand{\scalefigure}{0.3}
\begin{figure*}[htbp]
	\centering
	\begin{subfigure}[b]{.38\linewidth}
		\centering
		\includegraphics[scale=\scalefigure]{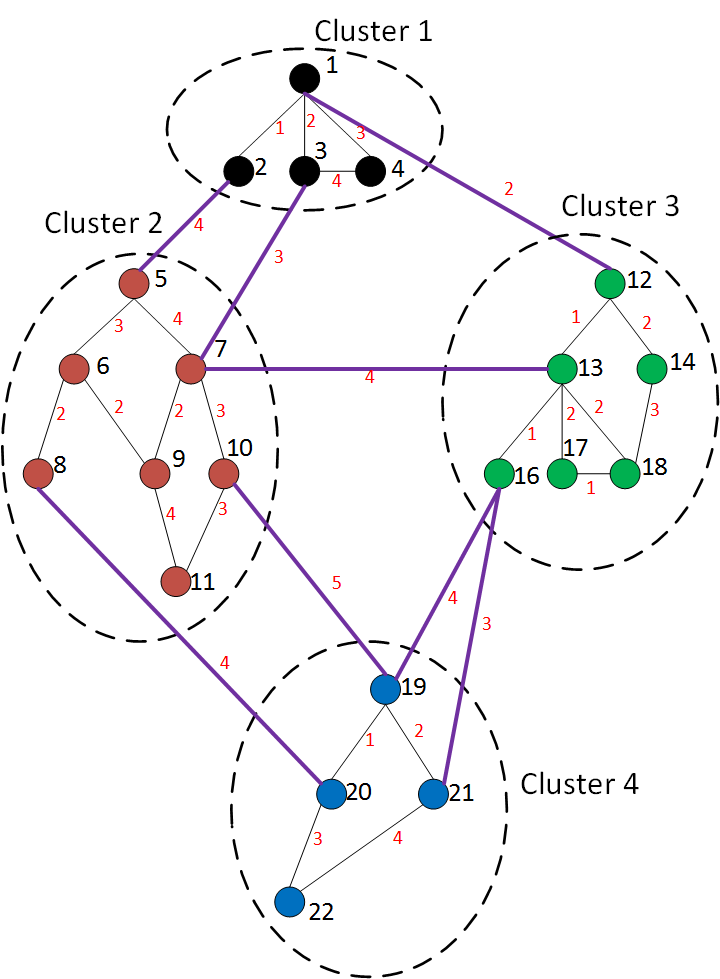}
		\caption{}
		\label{fig:fig2-a}
	\end{subfigure}
	\begin{subfigure}[b]{.2\linewidth}
		\centering
		\includegraphics[scale=\scalefigure]{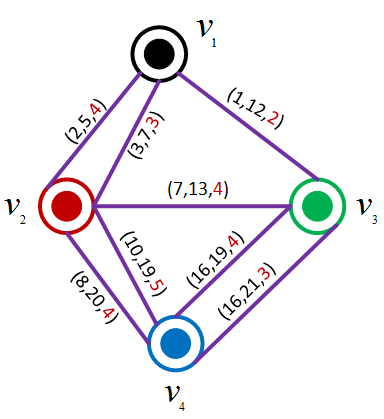}
		\textcolor{white}{\rule{6ex}{6ex}}
		\includegraphics[scale=\scalefigure]{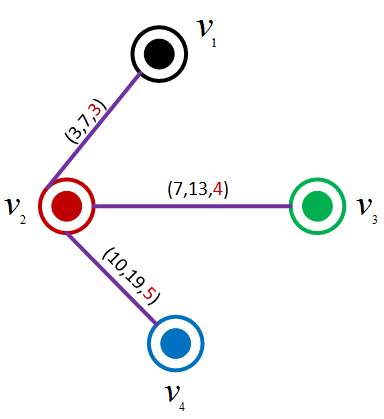}
		\caption{}
		\label{fig:fig2-b}
	\end{subfigure}
	\begin{subfigure}[b]{.38\linewidth}
		\centering
		\includegraphics[scale=\scalefigure]{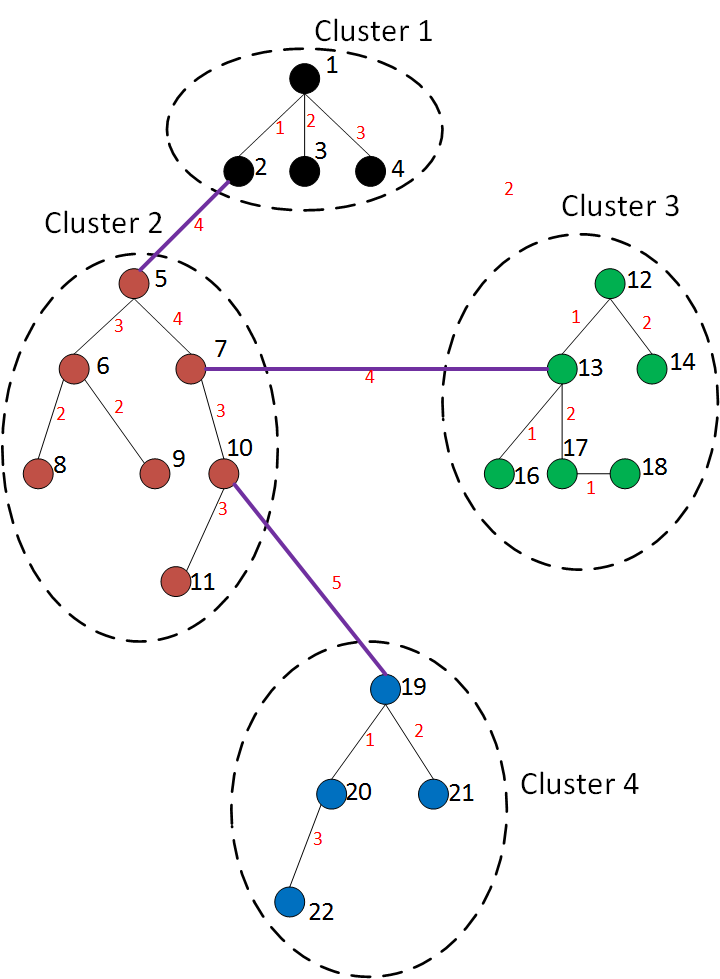}
		\caption{}
		\label{fig:fig2-d}
	\end{subfigure}
	\caption{A sample of creating a spanning cluster tree}
	\label{fig:fig_02}
\end{figure*}
\setlength{\intextsep}{0pt}

With reference to a graph given by Figure~\ref{fig:fig2-a}, denoted as $G$, the cluster graph $G'$ is given by Figure~\ref{fig:fig2-b}. All spanning trees of $G'$  are entire feasible combinations of inter-cluster edges. An arbitrary spanning tree $ST$ of $G'$ is given by Figure~\ref{fig:fig2-b}, this tree has three edges (3, 7, 3), (10, 19, 5), (7, 13, 4). Using Depth First Search to explore the tree, starting at $v_1$, the roots of all local trees are \{1, 7, 13, 19\}. Finally, finding all shortest path trees at these roots, an optimal cluster tree of $G$ with the inter-cluster edges combination $st$ can be found in Figure \ref{fig:fig2-d}.

Base on the upper procedure of construction an optimal cluster tree from a combination of cluster edges, this paper proposes a genetic algorithm to solve \gls{cluspt} where search space is all spanning trees of the cluster graph which can be built from given graph. This algorithm is called GACSPT. \\

\subsection{Individual representation}
The most significant determinant to the success or failure of a genetic algorithm is the representation of candidate solutions with the genetic operator applied to them. In studies, there are various methods for encoding spanning trees, such as the predecessor coding, Prufer numbers, link-and-node biasing, network random keys, edge-set and other representations which are used less often \cite{raidl2003edge, franz2006representations}. The previous studies have shown that encoding spanning tree by edge-set brings good performance for \gls{ga} \cite{raidl2003edge}. It is for this reason that, in this paper, edge-set structure is used to represent candidate solutions for \gls{cluspt}.

\begin{figure}[htbp]
	\centering
	\includegraphics[scale=0.25]{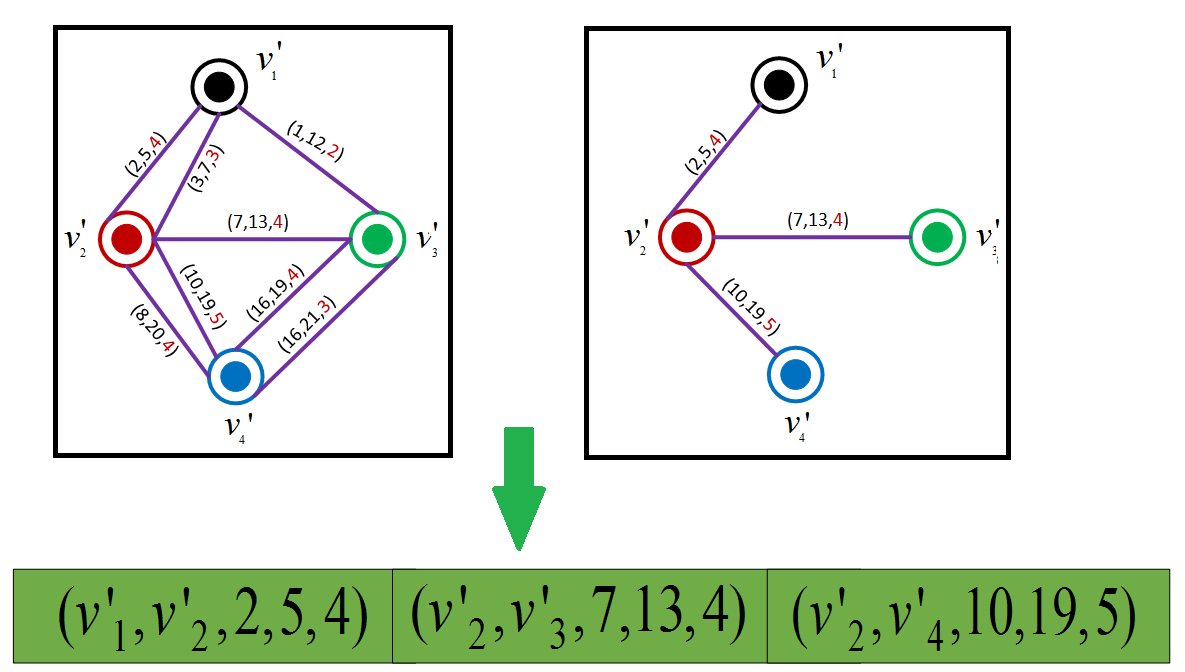}
	\caption{Example for individual representation.}
	\label{fig:representation}
\end{figure}

The information of two nodes connecting two clusters also have to be saved, consequently, a gene in a chromosome is to be formatted as ($v'_i,v'_j,u,v,c$); $v'_i,v'_j (i \neq j,i \le j \le k)$, which are clusters' notations; $u,v$ are vertices in $G$, $c$  is the cost of the edge between $u$ and $v$. An example for this representation is presented in Figure~\ref{fig:representation}.

\subsection{Population Initialization}
 Genetic algorithms fail to find a good solution if it falls into the local point of the search space. A random initialization plays an important role to make the search process investigate many different areas in the search space. To create a random spanning tree from a given graph, this paper uses the Prim random spanning tree algorithm (PrimRST) introduced in \cite{raidl2003edge}. This algorithm mechanism is equivalent to the Prim algorithm, however, in place of choosing the closest vertex, it chooses an arbitrary vertex and a random edge connects it to current spanning tree. For initializing a population with $N$ individuals, the procedure PrimRST is repeated $N$ times. PrimRST algorithm can be refered to Algorithm~\ref{alg:PrimRST}. PrimRST is a simple algorithm and easy for implementing, however it can create many different trees with distinct structures, therefore, it can give our algorithm a good start. Moreover, PrimRST has a small time complexity. 

\subsection{Crossover Operator}
\begin{figure}[htbp]
	\centering
	\includegraphics[scale=0.25]{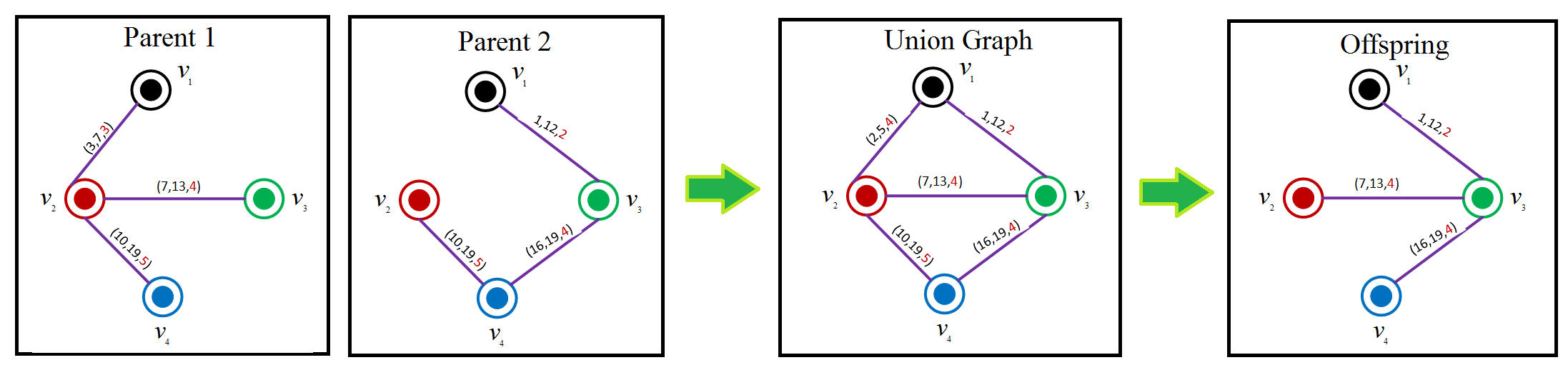}
	\caption{Steps of novel crossover operator.}
	\label{fig:crossover_operator_fig_04}
\end{figure}

Recombination plays an important role in genetic algorithm. A crossover operator is judged as a good operator if it can create an offspring which inherits good characteristics from parents.  Consequently, for \gls{cluspt}, recombination should build an offspring spanning tree that contains mostly or entirely of edges from parents. Therefore, this paper uses a crossover operator described in \cite{raidl2003edge} to create children from given parents. Denoting parents as $T_1, T_2$, the offspring $T$ can be constructed by getting a random tree from the graph $G_T =  (V,E(T_1) \cup E(T_2))$. Figure~\ref{fig:crossover_operator_fig_04} illustrate mechanism of the crossover operator in which the input graph is in Figure~\ref{fig:fig2-a}.

\begin{algorithm}[htbp]
	\KwIn{$G' = (V', E')$}
	\KwOut{A random spanning tree $T'$ of $G'$}
	\BlankLine
	\Begin
	{
		$s \leftarrow$ A random vertex in $V$\;
		$V(T') \leftarrow \{s\}$\;
		$E(T') \leftarrow \emptyset$\;
		$A \leftarrow \{e = (s,v) \in E', \forall v \in V'\}$\;
		\While{$|V(T')| < |V'|$}
		{
			$e' = (u,v) \leftarrow$ a random edge in $A$\;
			$A \leftarrow A \backslash \{e'\}$\;
			\If{$v \notin V(T')$}
			{
				$V(T') \leftarrow V(T') \cup \{v\}$\;
				$E(T') \leftarrow E(T') \cup \{e'\}$\;
				$A \leftarrow A \cup \{e = (v,p) \in E', \forall p \in V'\backslash V(T'))\}$\;
			}	
		}
	}
	\caption{PrimRST(V,E)}
	\label{alg:PrimRST}
\end{algorithm}

\subsection{Mutation Operator}
Mutation is one of the decisive operators in genetic algorithm. Mutation operator aids the search process not to fall in local optimum and possibly go to the global optimum. With observation from the problem and above discussion, inference can be made that the way one chooses inter-cluster edges influences significantly to the received cluster tree. Therefore, if another edge is replaced with one edge between two clusters in current combination of inter-cluster edges, the effectiveness of this combination would be affected.

\begin{figure}[htbp]
	\centering
	\includegraphics[scale=0.25]{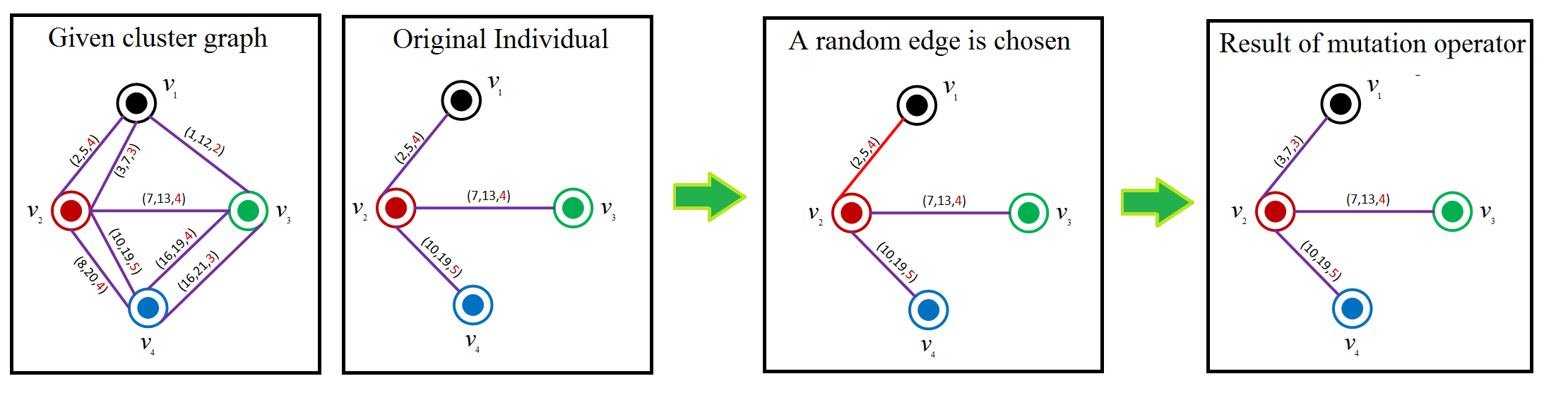}
	\caption{Steps of novel mutation operator.}
	\label{fig_05}
\end{figure}

Furthermore, cluster graph is a multi-graph and there are various ways to choose the edge between two clusters. As such, this paper proposes a new mutation operator for genetic algorithm to solve \gls{cluspt}. The novelty of this mutation operator starts by selecting a random edge in the current spanning tree. This random edge satisfies that there are more than one edge connecting two ends of it in the given cluster graph. After which, this random edge is replaced by another arbitrary edge between this two ends. This process can be clarified by Figure \ref{fig_05}.

\section{A new model optimization for \glsentryshort{cluspt}}
\label{sec_model}
In science and engineering, there are several optimization problems which can be modified as single-objective \gls{blop}. The formulation of \gls{blop} is defined that one optimization task nests another optimization task. More specifically, there is a pair of objective functions, namely $F: R^u \times R^l \rightarrow R$ and $f: R^u \times R^l \rightarrow R$. This two functions have a relationship as follows:
$$ Minimize_{x_u \in X_u, x_l \in X_l} F(x_u, x_l)$$
Subject to $\displaystyle x_l \in argmin\left\{f(x_u, x_l)\right\}$

In the above expression, $F$ is called the leader function and $f$ represents the follower function. Every time, the leader function makes a decision by choosing a specific vector $x_u \in X_u$ which $X_u \subseteq R^u$ is the search space of $F$. Thereafter, the follower function selects a candidate $x_l$ which $X_l \subset R^l$ from the design space $X_l \subseteq R^l$, such that $f$ is optimized given the leader’s preceding decision.

Previous algorithm selects a combination of all inter-cluster edges first and proceeds to determines all the roots of the local trees to construct a solution. Now, consider a combination of all the roots of the local trees first and based on this combination, we find all inter-cluster edges. Let $G = (V, E, w)$ be a graph, $C = \left\{C_1, C_2,\ldots, C_k\right\}$ be a partition of $V$, $r$ be the root of $G$. With no loss of generality, we assume that $r \in C_1$.

Consider a combination $U = \left\{r_1, r_2,\ldots, r_k\right\}$ which $r_i \in C_i, \forall i = 1,\ldots, k$ and $r_1 = r$ ($r$ is the root of $G$, so $r$ is always the root of the local tree of $C_1$). A cluster tree which has $r_1, r_2,\ldots, r_k$ as the roots of local trees is built as following:
\begin{itemize}
	\item First step: Create a unweighted directed graph $H = (V_H, E_H)$
	\item Second step: Select a directed tree $L$ of $H$ which a solution for \gls{cluspt} based on $U$ and $L$ can be found to optimize the objective
\end{itemize}

In which, the cluster graph $H = (V_H, E_H)$ is constructed as following:
\begin{itemize}
	\item Each vertex in $V_H$ stands for a cluster in $V$ and $v_i$ stands for $C_i(1 \leq i \leq k)$.
	\item If there is one edge $e = (u, r_i) \in E$ which $u \in C_j$ and $r_i \in C_i$ is the vertex, chosen to become the root of the local tree of cluster $C_i$ then a directed edge $e = (v_j, v_i)$ is joined to the edge set $E_H$.
\end{itemize}
The Figure~\ref{fig:fig6} illustrates above steps. 

\setlength{\intextsep}{0pt}
\renewcommand{\scalefigure}{0.2}

\begin{figure}[htbp]
	\centering
	\begin{subfigure}[b]{.28\linewidth}
		\centering
		\includegraphics[scale=\scalefigure]{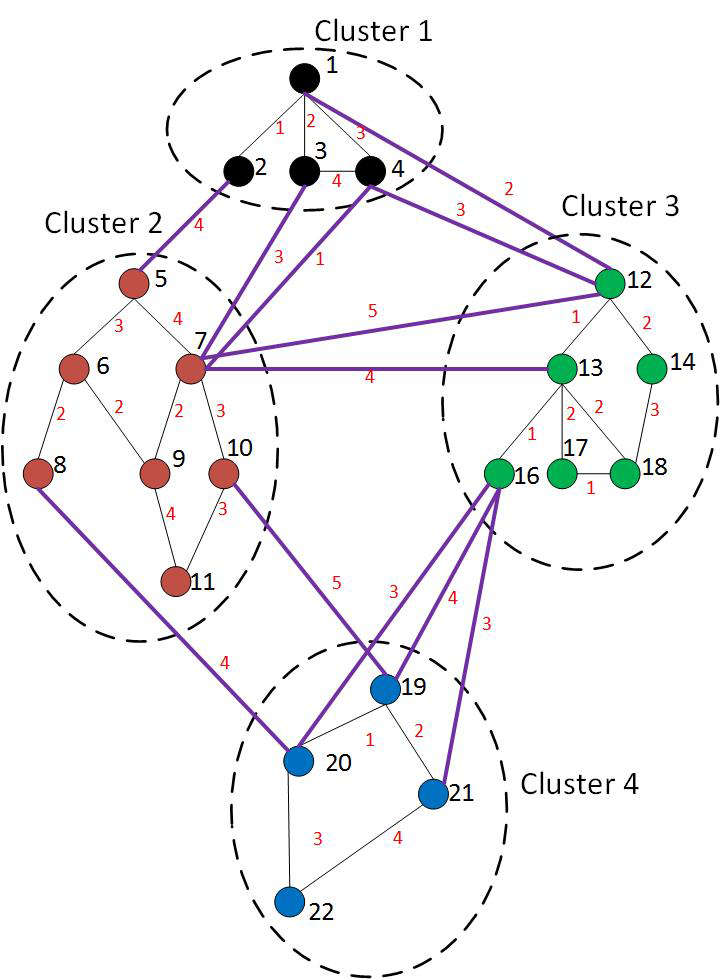}
		\caption{}
		\label{fig:fig6-a}
	\end{subfigure}
	\begin{subfigure}[b]{.28\linewidth}
		\centering
		\includegraphics[scale=\scalefigure]{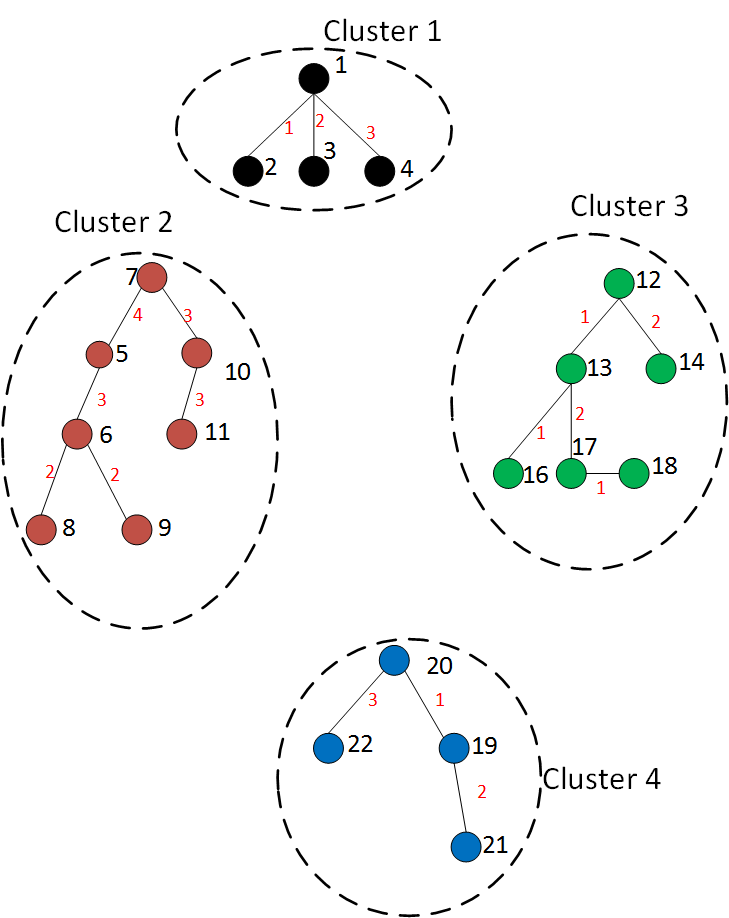}
		\caption{}
		\label{fig:fig6-b}
	\end{subfigure}
	\begin{subfigure}[b]{.14\linewidth}
		\centering
		\includegraphics[scale=\scalefigure]{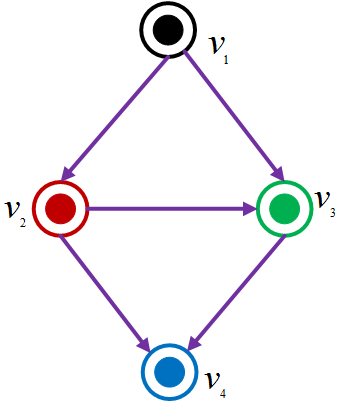}
		\textcolor{white}{\rule{3ex}{3ex}}
		\includegraphics[scale=\scalefigure]{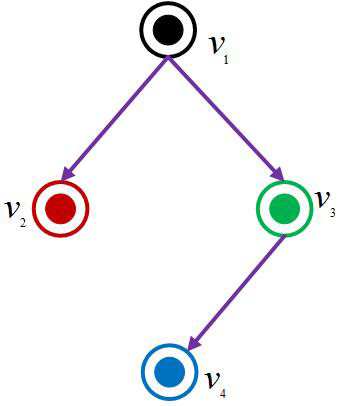}
		\caption{}
		\label{fig:fig6-d}
	\end{subfigure}
	\begin{subfigure}[b]{.28\linewidth}
		\centering
		\includegraphics[scale=\scalefigure]{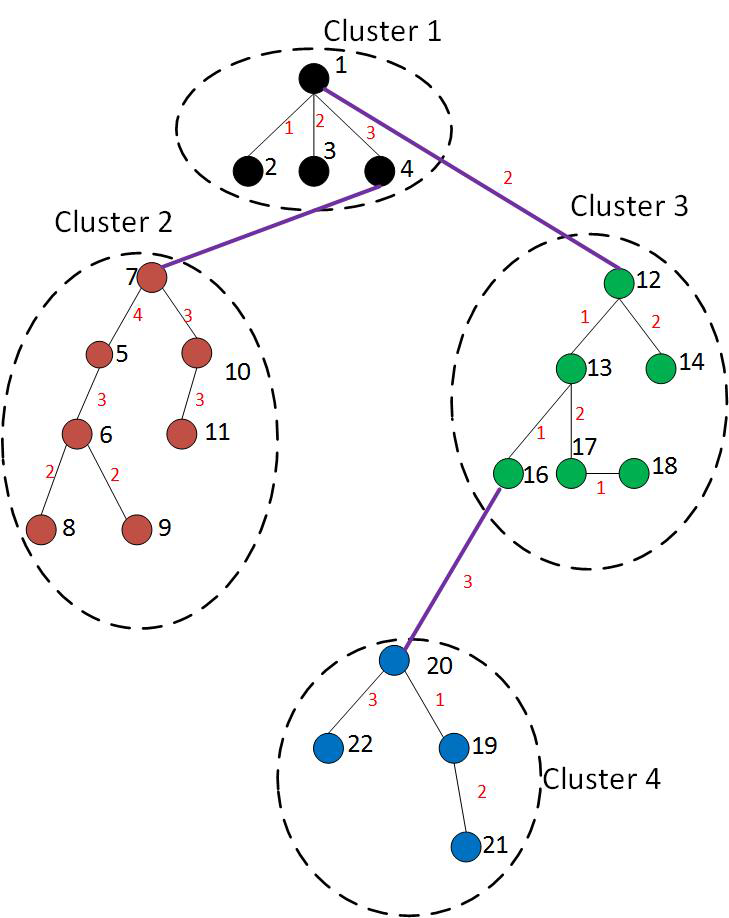}
		\caption{}
		\label{fig:fig6-e}
	\end{subfigure}
	\caption{Steps of creating a clustered spanning tree based on roots combination}
	\label{fig:fig6}
\end{figure}
\setlength{\intextsep}{0pt}

Based on these analyses, the \gls{cluspt} also can be modified and transformed to a model that is similar to bi-level optimization problem. This model is defined as follows:

Let $X_u$ be the set, which includes all combinations of roots of local trees. With a combination $u \in X_u$, denote $H_u$  is the directed graph built base on $u$ (the method to build this graph is mentioned above). $S(H_u)$ is the set which consists of all directed trees of $H_u$. With $u \in X_u$ and $v \in S(H_u)$, a cluster tree or a solution of \gls{cluspt} can be built. Denote $F(u,v)$ is the value of the objective.
\[
Minimize_{x_u \in X_u, x_l \in S(H_u)} F(x_u,x_l) 
\]

The similarity between bi-level optimization problem and this formulation is that every time, the leader makes a decision by choosing a combination $x_u \in X_u$. Thereafter, the follower function also has to find a candidate $x_l \in X_l$ to minimize the given function $F(x_u,x_l)$. The difference is that in this problem, the search space of the follower function is built base on $x_u$ and we only need to minimize the follower function. Despite of those differences, the previous approaches proposed for bi-level optimization problem could be used to solve \gls{cluspt}. One of those techniques is using Nested Bi-Level Evolutionary Algorithm (N-BLEA). The main feature of this approach is inspired by the nature of \gls{blop} which is that the lower function being tackled with respect to each population member in the upper level. If this algorithm is applied for \gls{cluspt}, we would be challenged with time consumption because this algorithm has to run evolutionary algorithm many times to find the optimum for follower functions. 

This paper changes the algorithm N-BLEA on the upper level to be applicable with our problem. To replace the EA for the leader function, a heuristic algorithm can be used, specifically local search (N-LSEA). The algorithm starts with an initial candidate $x_u$. $Neighbor(x_u)$ is the function which returns all feasible neighbors of  $x_u$. With respect to a candidate solution $x_{u,j}$ in the neighbor set, the corresponding lower level function takes the form $F(x_{u,j},x_l)$ and the search space $S(H_{x_{u,j}})$. The \gls{ga} is employed to find the optimal solution $x_{l,j}^*$ in this search space for this level. Thereafter, evaluate $x_{u,j}$ by $F(x_{u,j}, x_{l,j}^*)$. After the above steps, a fittest candidate in the neighbor set can be acquired to start the search again. For a summary of N-LSEA, refer to Algorithm~\ref{alg:Nested Local Search Evolutionary Algorithm (N-LSEA)}.

$Neighbor(x_u)$ includes all combinations $x_v$ which differs from $x_u$ at exact one position. Assume that $x_u=\{r^{u}_{1},r^{u}_{2}, \ldots,r^{u}_{k-1},r^{u}_{k}\}$ such that $r^{u}_{i} \in C_i$  $\forall i=1 \ldots k$ and $r_1^u = r$, $x_v$ is neighbor of $x_u$ if and only if $x_v$ has format $x_v=\{r^{v}_{1}, r^{v}_{2}, \ldots, r^{v}_{k-1}, r^{v}_{k}\}$ which  $\exists j \neq 1: r^{v}_{j} \neq r^{v}_{j}$, and $\forall i \neq j: r^{u}_{j}= r^{v}_{j}$. The \gls{ga} which applied to search solutions for lower level (GAFLL) is similar to GACSPT. To represent individual, GAFLL use edges representation. To create an arbitrary directed tree from a single directed graph, GAFLL also uses PrimRST algorithm, however, as compared with GACSPT, this algorithm changes slightly to be applicable with single directed graph. Replace of starting by an arbitrary vertex as in GACSPT, GAFLL start searching with the root of cluster graph. The crossover operation for GAFLL is similar to the one for GACSPT. The offspring is generated by PrimRST algorithm from the graph which is the union of two parents.

\begin{algorithm}[htbp]
	\BlankLine
	\Begin
	{	
		$x_u \leftarrow$ A Random combination of roots of local trees\;
		$S(H_{x_u}) \leftarrow$ The search space for lower level of $x_u$\;
		\tcc{Find optimization solution of lower level by performing GAFLL on $S(H_{x_u})$}
		$x_l^* \leftarrow$ Apply GAFLL on $S(H_{x_u})$\; 
		Evaluate $x_u$ as $F(x_u, x_l^*)$\;
		T $\leftarrow$ 0\;
		\While{stop condition are not satisfied}
		{
			$N_{T,u} \leftarrow$ Neighbour($x_u$)\;
			\ForEach{$x_{u,j}$ in $N_{T,u}$}
			{
				$S(H_{x_{u,j}}) \leftarrow$ The search space for lower level of $x_{u,j}$\;
				$x_{l,j}^* \leftarrow$ Apply GAFLL on $S(H_{x_{u,j}})$\; 
				Evaluate $x_{u,j}$ as $F(x_{u,j}, x_{l,j}^*)$\;
				\If{$F(x_{u,j}, x_{l,j}^*) < F(x_u,x_l^*)$}
				{
					$x_u \leftarrow x_{u,j}$
				}
			}
			T $\leftarrow$ T + 1\;
		}
	}
	\caption{Nested Local Search Evolutionary Algorithm (N-LSEA)}
	\label{alg:Nested Local Search Evolutionary Algorithm (N-LSEA)}
\end{algorithm}

Because the search space for lower level is created from single directed graph, GAFLL has to use a new mutation operator. The novelty of this operator starts by selecting a directed edge $e = (v_i, v_j)$ in the cluster graph. This edge is not in the current individual. This individual always has a directed edge which has a tail $v_j$, denoting this edge as $e' = (v_i, v_j)$. The mutation operator replaces $e'$ by $e$. This process can be clarified by Figure~\ref{fig_7}.

\renewcommand{\scalefigure}{0.33}
\begin{figure*}[htbp]
	\centering	
	\setlength\tabcolsep{0 pt}
	\includegraphics[scale=\scalefigure]{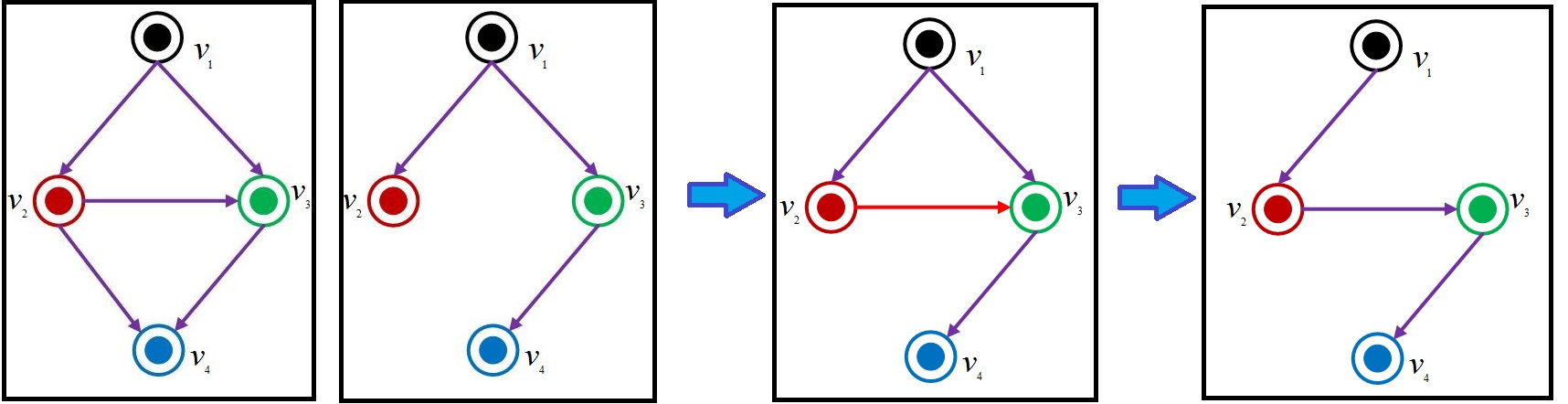}
	\centering
	\caption{Steps of novel mutation operator for multitasking evolutionary algorithm}
	\label{fig_7}
\end{figure*}

As highlighted by Algorithm~\ref{alg:Nested Local Search Evolutionary Algorithm (N-LSEA)}, multiple lower level tasks are solved when a candidate in upper level is considered. Besides, two arbitrary neighbors of one candidate are almost similar. Thus, it can be clear that evolutionary multitasking can be taken into consideration and applied to improve GAFLL.

Consider a candidate $u$ in the upper level, denote $S$ as the set of all feasible neighbors of $u$. To apply evolutionary multitasking for all lower levels of those neighbors (M-FLL), $S$ is firstly divided into clusters. Assume that all neighbors of $u$ are clustered into a set of $m$ groups denoted as $P = \{p_1, p_2,...,p_m\}$. The number of clusters of $P$ is determined by the sizes of these clusters. To be applicable with the problem, this algorithm divides all neighbors into groups each of which has at most two candidates. To increase the probability of fruitful information transfer during the multitasking process, two candidates having exactly one different position are combined. A summary of all steps for multitasking in nested local search evolutionary algorithm (M-LSEA) is provided in Algorithm~\ref{alg:Multitasking Local Search Evolutionary Algorithm (M-LSEA)}.

In Algorithm~\ref{alg:Multitasking Local Search Evolutionary Algorithm (M-LSEA)}, with regard to a group having two candidates $(u,v)$,  it can be determined that this algorithm has to find the search spaces of those candidates. The search spaces of those two can be different. $H_u$ and $H_v$ denote the cluster graphs of $u$ and $v$. Because $u$ differs from $v$ at exact one position, denoted by position $q$, the graph $H_u$  and $H_v$ have the same vertex set and their edge set are almost similar, except for all the edges whose tails are $v_q$.

\begin{figure}[htbp]
	\centering	
	\includegraphics[scale=1]{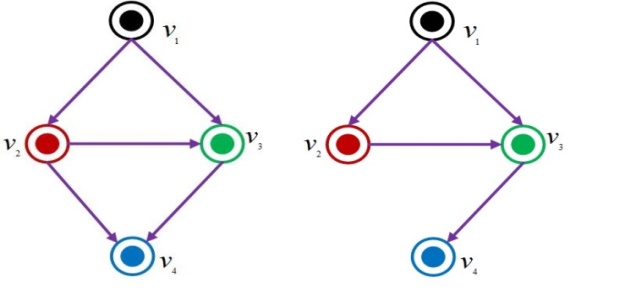}
	\centering
	\caption{The clusters graph $H_u$ and $H_v$ which be built based on $u=\{1,7,12,20\}$ and $v =\{1,7,12,21\}$}
	\label{fig:fig_8}
\end{figure}

For example, consider the given graph in Figure~\ref{fig:fig6-a} and two different combinations of roots of local trees are $u=\{1,7,12,20\}$ and $v=\{1,7,12,21\}$ . The cluster graphs $H_u$ and $H_v$ are given by Figure~\ref{fig:fig_8}. $u$ differs from $v$ at the $4^{th}$ position, hence, $H_u$ differs from $H_v$ in the edges connected to $v_4$. The edges connecting to $v_4$ of $H_u$ are ${(v_2,v_4),(v_3,v_4)}$ while that of $H_v$ is $\{(v_3,v_4)\}$.

\begin{algorithm}[htbp]
	\BlankLine
	\Begin
	{	
		$x_u \leftarrow$ A Random combination of roots of local trees\;
		$S(H_{x_u}) \leftarrow$ The search space for lower level of $x_u$\;
		
		\tcc{Find optimization solution of lower level by performing GAFLL on $S(H_{x_u})$}
		$x_l^* \leftarrow$ Apply GAFLL on $S(H_{x_u})$\; 
				
		Evaluate $x_u$ as $F(x_u, x_l^*)$\;
		T $\leftarrow$ 0\;
		\While{stop conditions are not satisfied}
		{
			$N_{T,u} \leftarrow$ Neighbour($x_u$)\;
			Cluster $N_{T,u}$ into a set of groups P\;
			\ForEach{group $p_i$ in P}
			{
				With regard to every candidate in $p_i$, create the search spaces for lower functions.\;
				Solve lower level optimization tasks for all $c_j \in p_i$ together via MFEA\;
				\ForEach{$c_j \in p_i$}
				{
					Evaluate $c_j$ as $F(x_{u,j}, x_{l,j}^*)$\;
					\If{$F(x_{u,j}, x_{l,j}^*)<F(x_u,x_l^*)$}
					{
						$x_u \leftarrow x_{u,j}$
					}
				}
			}
			T $\leftarrow$ T + 1\;
		}
	}
	\caption{Multitasking Local Search Evolutionary Algorithm (M-LSEA)}
	\label{alg:Multitasking Local Search Evolutionary Algorithm (M-LSEA)}
\end{algorithm}

Because of the difference between two search spaces, the individual created for one lower task may not be feasible for the others. Because the difference between two search spaces, the individual created for one lower task may not be feasible for the other. However, when running multitasking evolutionary algorithm, one always choose an individual to the feasible task by determining its skill factor. Initially, if a rooted tree is created based on a cluster graph of one task, then its skill factor is set by this task. When applying assortative mating to a parent pair, the offspring generated is always a feasible rooted tree for one task or both tasks. This proposition can be easily figured out from the difference between two cluster graphs of two tasks as mentioned above. Due to this reason, in vertical cultural transmission, if an offspring is applicable for only one task, its skill factor is set by this task. On the other hand, if it is feasible for both tasks, the algorithm follows the original multitasking evolutionary algorithm. The details of assortative mating and vertical cultural transmission for M-LSEA are given in Algorithm~\ref{alg:Assortative mating} and Algorithm~\ref{alg:Vertical cultural transmission via selective imitation}. \\

\begin{algorithm}[htb]
	\BlankLine
	\Begin
	{	
		Consider two individuals which are chosen from current-pop\;
		Generate a random number rand between 0 and 1\;
		\eIf{($\tau_1= \tau_2$) or ($rand < rmp$)}
		{
			Crossover $p_1$ and $p_2$ to generate a new offspring c\;
		}
		{
			Mutate parent $p_1$ to create offspring $c_1$\;
			Mutate parent $p_2$ to create offspring $c_2$\;			
		}
	}
	\caption{Assortative mating}
	\label{alg:Assortative mating}
\end{algorithm}


\begin{algorithm}[htbp]
	\BlankLine
	\Begin
	{	
		An offspring, denoted as c, will have two parents ($p_1$ and $p_2$) or have single parent ($p_1$ or $p_2$) (Refer Algorithm 4)\;
		\eIf{c has two parents}
		{
			\eIf{c is feasible for both tasks}
			{
				$rand \leftarrow$ a random number between 0 and 1\;
				\eIf{$rand < 0.5$}
				{
					c imitates $p_1$ (the offspring is evaluated only for task $\tau_1$, the skill factor of $p_1$)\;
				}
				{
					c imitates $p_2$ (the offspring is evaluated only for task $\tau_2$, the skill factor of $p_2$)\;
				}
			}
			{
				\eIf{c is feasible for task $\tau_1$ (c is rooted tree of the cluster graph of task $\tau_1$)}
				{
					c imitates $p_1$\;
				}
				{
					c imitates $p_2$\;
				}
			}
		}
		{
			c imitates its single parent\;
		}
		Factorial cost of c with respect to the task which is not its skill factor is set equal to $\infty$.\;
	}
	\caption{Vertical cultural transmission via selective imitation}
	\label{alg:Vertical cultural transmission via selective imitation}
\end{algorithm}


\section{Computational results}
\label{sec_results}

\subsection{Problem instances}
This section presents the experimental results of three algorithms we mentioned to solve \gls{cluspt} in previous sections.  Due to the fact that there are no available set of instances of the \gls{cluspt}, we generated a set of test instances based on the MOM~\cite{helsgaun_solving_2011, mestria_grasp_2013} (further on in this paper MOM-lib) of the Clustered Traveling Salesman Problem. The MOM-lib consists of six distinct types of instances which were created through various algorithms~\cite{demidio_clustered_2016} and categorized into two kinds according to the dimension: small instances, each of which has between 30 and 120 vertices, and large instances, each of which has over 262 vertices. The instances are suitable for evaluating clustered problems~\cite{mestria_grasp_2013}. All instances of the \gls{cluspt} problem are available via \cite{Pham_Dinh_Thanh_2018_Instances}.


\subsection{Experimental setup}
To evaluate the performance of three algorithms GACSPT, \gls{n-lsea} and \gls{m-lsea}, we implement all algorithms in same environment (Intel Corei7 – 4790M - 3.60GHz, 16GB). We compare our algorithms as follows:

\begin{itemize}
    \item In the first experiment, we compare the performance of 4 new algorithms: GACSPT, \gls{n-lsea},\gls{m-lsea} and EAM on all instances in terms of solution quality and running time.
    \item In the second experiment, since the performance of the proposed algorithms were contributed by parameter: number of clusters in the \gls{cluspt} instance and the average number of vertices in a cluster, the convergence trends of each task in generations, we conducted experiments for evaluating the effect of these parameters.    
\end{itemize}

All results for each algorithm presented in this section are summaries of 30 independent runs for each instance. During each run, the genetic algorithm GACSPT runs with a population with 200 individuals, this population is evolved for a maximum of 6000 generations or until the solutions converge to the optimum. For \gls{n-lsea} and \gls{m-lsea}, the upper level is simulated until the solutions reach the optimum, the lower levels runs a population with 20 individuals. When running one task in \gls{n-lsea}, and running two tasks in \gls{m-lsea}, we always set total the number of task evaluation to be 1200. The random mating probability is 0.9 and mutation rate is set to 0.1 in all experiments. 

\subsection{Experimental results}
We use the following criteria to measure the output’s quality of the algorithms.

\begin{center}
    \begin{tabular}{p{4cm} p{10.5cm}} 
    	\hline
        \multicolumn{2}{c}{\textbf{Criteria}} \\
        \hline
        \emph{Avg} & The averaged value after 30 runs for each instance. \\                  
        \addlinespace
        \emph{Coefficient of variation (CV)} & Ratio of standard deviation to the average function value. \\      
        \addlinespace       
        \emph{Number of Better Instances (NIB)} & Number of instances in which results obtained by Multi-tasks are better than those obtained by Single task.\\                              
        \addlinespace
        \emph{NAB} & Numerical instances in which single task outperform multi-tasks. \\
       	\addlinespace          
        \emph{BF} & Best function value achieved over all runs. \\
        \addlinespace
        \emph{Rm} & Running time of algorithms in minutes.\\
        \hline
    \end{tabular}
\end{center}

To compare the performances of two algorithms, we compute the \gls{rpd}~\cite{pop_two-level_2018} between the average results obtained by the algorithms. The \gls{rpd} is computed by the following formula:
\begin{equation}
	RPD = \dfrac{Solution - Best}{Best}*100\%
\end{equation}
where $Solution$ is solution provided by each of algorithms:
GACSPT, \gls{m-lsea} and \gls{n-lsea} and $Best$ is the solution obtained by EAM.

We also compute the gap between the costs of the results obtained by algorithm A and B with the following formula:
\begin{equation}
	PI(A,B)=\dfrac{C_B - C_A}{C_B}*100\%
\end{equation}
where $C_A$, $C_B$ is the cost of the best solution generated by algorithm A, B respectively.

\subsubsection{Non-parametric statistic for comparing the results of the proposed algorithms}
\label{sec:results:subsec:NonParamtricStatistic}
In this section, we study the results obtained by proposed algorithms on the use of the non-parametric tests. Our study has two main steps:
\begin{itemize}
	\item The first step first uses the statistical methods i.e. Friedman, Aligned Friedman, Quade~\cite{derrac_practical_2011, carrasco_recent_2020} for testing the differences among the related samples means, that is, the results obtained by each algorithm.
	\item The second step is performed when the test in the first step rejects the hypothesis of equivalence of means, the detection of the concrete differences among the algorithms can be done with the application of post-hoc statistical procedures~\cite{derrac_practical_2011, carrasco_recent_2020}, which are methods used for comparing a control algorithm with remain algorithms.
\end{itemize}

Tables~\ref{tab:Results_Type2_3Algorithms}
-\ref{tab:Results_Type6Large_3Algorithms} summarize the results obtained in the competition organized by types of instances and three algorithms.

\begin{table}
	\centering
	\caption{Results of the Friedman and Iman-Davenport test ($\alpha$=0.05)}\label{tab:Results_Friedman}
	\begin{tabular}{c c c c c c}
		\toprule
		Friedman Value & Value in $X^2$ & $p$-value & Iman-Davenport Value & Value in $F_F$ & $p$-value \\
		\cmidrule(l{3pt}r{3pt}){1-3} 
		\cmidrule(l{3pt}r{3pt}){4-6} 
		\textbf{41.355} & 5.991 & 1.085E-9 & \textbf{25.807} & 3.042 & 1.097E-10\\
		\bottomrule
	\end{tabular}
\end{table}

Table~\ref{tab:Results_Friedman} shows the result of applying Friedman's and Iman-Davenport's tests. Given that the statistics of Friedman and Iman-Davenport are greater than their associated critical values, there are significant differences among the observed results with a probability error $p \leq 0.05$.

\setlength{\intextsep}{6pt}
\begin{table}
	\centering
	\caption{Average rankings achiedved by the Friedman, Friednman Aligned, and Quade tests}\label{tab:Results_Rank}
\begin{tabular}{c c c c}
	\toprule
	Algorithms & Friedman & Friednman Aligned & Quade \\
	\midrule
	GACSPT & 1.730 & 105.690 & 1.502 \\
	
	\glsentrytext{m-lsea} & 1.745 & 154.245 & 1.770 \\
	
	 \glsentrytext{n-lsea} & 2.525 & 191.565 & 2.728 \\
	\bottomrule
\end{tabular}
\end{table}

\setlength{\intextsep}{6pt}

Table~\ref{tab:Results_Rank} summarizes the ranking obtained by the Friedman, Friednman Aligned, and Quade tests. Results in Table~\ref{tab:Results_Rank} strongly suggest the existence of significant differences among the algorithms considered.

The results on Table~\ref{tab:Results_Rank} show that the rank of GACSPT algorithm is smallest, so we choose GACSPT as the control algorithm. After that, we will apply more powerful procedures, such as Holm's and Holland's etc., for comparing the control algorithm with the rest of algorithms. Table~\ref{tab:CompareControlAlg} shows all the possible hypotheses of comparison between the control algorithm and the remaining, ordered by their $p$-value and associated with their level of significance 

\setlength{\intextsep}{6pt}
\begin{table}[!htp]
\centering
\caption{The z-values and p-values of the Friedman Aliged, Quade procedures (GACSPT is the control algorithm)} \label{tab:CompareControlAlg}
\begin{tabular}{cc cccc cccc}
\toprule
& & \multicolumn{4}{c}{\textbf{Friedman Aliged}} &
\multicolumn{4}{c}{\textbf{Quade}}\\

\cmidrule(l{3pt}r{3pt}){1-2}
\cmidrule(l{3pt}r{3pt}){3-6}
\cmidrule(l{3pt}r{3pt}){7-10} 
$i$ & 
algorithms &
$z$ &
$p$ &
Holm &
Holland &
$z$ &
$p$ &
Holm &
Holland\\

\midrule
2 & NEST & 7.000 & 2.559E-12 & 0.025 & 0.025 & 7.529 & 5.113E-14 & 0.025 & 0.0253\\
1 & MFEA & 3.957 & 7.561E-5 & 0.05 & 0.050 & 1.650 & 0.099 & 0.05 & 0.050\\
\bottomrule
\end{tabular}
\end{table}

\setlength{\intextsep}{6pt}

Table~\ref{tab:Adjust_p_FriedmanAliged} and Table~\ref{tab:Adjust_p_QUADE} show all the adjusted $p$ values for each comparison which involves the control algorithm. The $p$ value is indicated in each comparison and we stress in bold the algorithms which are worse than the control algorithm, considering a level of significance $\alpha = 0.05$.

\setlength{\intextsep}{6pt}
\begin{table}[htbp]
\centering
\caption{Adjusted p-values for the Friedman Aligned test (GACSPT is the control method)} \label{tab:Adjust_p_FriedmanAliged}
\begin{tabular}{c c ccccc}
\toprule
i & Algorithms & Unadjusted $p$ & $p_{Holland}$ & $p_{Holm}$ & $p_{Hochberg}$ & $p_{Hommel}$\\
\midrule
1 & \textbf{\glsentrytext{n-lsea}} & 5.119E-12  & 2.559E-12 & 5.119E-12 & 5.119E-12 & 5.119E-12\\
2 & \textbf{\glsentrytext{m-lsea}} & 7.560E-5 & 1.512E-4 & 7.560E-5 & 7.560E-5 & 7.560E-5\\
\bottomrule
\end{tabular}
\end{table}


\setlength{\intextsep}{6pt}

\setlength{\intextsep}{6pt}

\begin{table}[htbp]
\centering
\caption{Adjusted p-values for the QUADE test (GACSPT is the control method)} \label{tab:Adjust_p_QUADE}
\begin{tabular}{ccccccc}
\toprule
i & Algorithms & Unadjusted $p$ & $p_{Holland}$ & $p_{Holm}$ & $p_{Hochberg}$ & $p_{Hommel}$\\
\midrule
1 & \textbf{\glsentrytext{n-lsea}} & 5.112E-14 & 1.021E-13 & 1.023E-13 & 1.023E-13 & 1.023E-13\\
2 & \glsentrytext{m-lsea} & 0.0989 & 0.099 & 0.099 & 0.099 & 0.0989\\
\bottomrule
\end{tabular}
\end{table}
%

\setlength{\intextsep}{6pt}

\subsubsection{Comparison the performances of proposed algorithms and existing algorithms}

As mention in subsection~\ref{sec:results:subsec:NonParamtricStatistic}, GACSPT has the lowest ranking and \gls{g-mfea}~\cite{thanh2020efficient} is one of the most effective algorithms for solving the \gls{cluspt}. Therefore, this subsection will compare \gls{g-mfea} with GACSPT by using the Wilcoxon signed rank test~\cite{derrac_practical_2011, carrasco_recent_2020} to detect differences in both means.

\begin{table}[htbp]
	\centering
	\caption{Wilcoxon signed ranks test results} \label{tab:wilcoxon}%
	\begin{tabular}{ c r r r r}
		\toprule
		\multicolumn{1}{r}{} & 
		\multicolumn{1}{c}{\textbf{N}} & 
		\multicolumn{1}{c}{\textbf{Mean Rank}} & 
		\multicolumn{1}{c}{\textbf{Sum of Ranks}} & 
		\multicolumn{1}{c}{$p$-\textbf{value}} \\
		
		\midrule
		$R^-$ & 15    & 22.93 & 344.0 & 7.48836E-05 \\

		$R^+$ & 43    & 31.79 & 1367.0 &  \\

		Ties  & 0     &       &       &  \\

		Total & 58    &       &       &  \\
		\bottomrule
	\end{tabular}%
\end{table}%

Table~\ref{tab:wilcoxon} presents the comparison between GACSPT and \gls{g-mfea}. As the table status, the $p$-value is less than 0.05 so the null hypothesis of equality of means is rejected. It is mean that a given algorithm overcomes the other one, with a level of significance $\alpha = 0.05$. Table~\ref{tab:wilcoxon} also points out that GACSPT shows a significant improvement over \gls{g-mfea}.

In the next subsections, more detail of comparison between GACSPT, \gls{m-lsea} and \gls{n-lsea} are presented. In which we focus on determine the impact of the number of clusters on the performance of those algorithms.  
\subsubsection{Comparison the performances of GACSPT and \gls{m-lsea}}
The running time of \gls{m-lsea} is shorter than GACSPT in almost all instances. The reason behind this result is that by using \gls{m-lsea}, with respect to the local search steps, the algorithm can converge to the optimum faster than genetic algorithm.

The experimental data shows that with regard to instances having small number of clusters, \gls{m-lsea} outperforms GACSPT. As we can figure out that the numbers of clusters of Type 2, Type 3 and Type 4 instances are smaller than 11 and Tables~\ref{tab:Results_Type2_3Algorithms}, \ref{tab:Results_Type3_3Algorithms}, \ref{tab:Results_Type4_3Algorithms} show that the results obtained by \gls{m-lsea} are better than GACSPT in all instances of these types. However, the value PI obtained from Euclidean instances is not too large. The max value PI is 2.35\% which belongs to the instance 4i400a, and the smallest is 0.015\%. 
We can see this performance in Figure~\ref{fig:11}. The biggest gap which \gls{m-lsea} outperforms GACSPT is 16.61 of the instances 6i300, and the instance 10C1k.0 gives the smallest gap, which is 0.171.
 
\begin{landscape}
\begin{table*}[htbp]
  \centering
  \caption{Results Obtained by GACSPT, N-LSEA and M-LSEA for Instances of Type 2}

    \begin{tabular}{l r r r r r r r r r}
    \toprule
    \multicolumn{1}{c}{\multirow{2}[4]{*}{\textbf{Instances}}} & 
    \multicolumn{3}{c}{\textbf{GACSPT}} &
    \multicolumn{3}{c}{\textbf{M-LSEA}} & 
    \multicolumn{3}{c}{\textbf{N-LSEA}} \\
    
	\cmidrule(l{3pt}r{3pt}){2-4} \cmidrule(l{3pt}r{3pt}){5-7}  \cmidrule(l{3pt}r{3pt}){8-10} 
	& 
	\multicolumn{1}{c}{\textbf{BF}} & 
	\multicolumn{1}{c}{\textbf{Avg}} & 
	\multicolumn{1}{c}{\textbf{Rm}} & 
	
	\multicolumn{1}{c}{\textbf{BF}} & 
	\multicolumn{1}{c}{\textbf{Avg}} & 
	\multicolumn{1}{c}{\textbf{Rm}} & 
	
	\multicolumn{1}{c}{\textbf{BF}} & 
	\multicolumn{1}{c}{\textbf{Avg}} & 
	\multicolumn{1}{c}{\textbf{Rm}} \\
	
	\midrule
    10C1k.0 & 603982229.7 & 604675739.3 & 13.8  & 603896769.1 & 604075680.5 & 7.9   & 603896769.1 & 604205712.3 & 7.9 \\
     
    10C1k.1 & 555719300.5 & 556214402.2 & 16.6  & 555519829.5 & 555601193 & 6.5   & 555519829.5 & 555655243.8 & 6.8 \\
     
    10C1k.2 & 740604616.5 & 741022869.8 & 15.6  & 740549053.6 & 740684380.1 & 7.9   & 740549053.6 & 740844678.6 & 8.1 \\
     
    10C1k.3 & 594441714.7 & 594586353.6 & 15.5  & 594412757.2 & 594496659.8 & 7.7   & 594412757.2 & 594664542.2 & 8.0 \\
     
    10C1k.4 & 533230914.9 & 533679392.5 & 12.2  & 533153746.9 & 533199581.9 & 7.8   & 533153746.9 & 533241310.4 & 8.0 \\
     \addlinespace
    10C1k.5 & 582566421.7 & 582865515.5 & 6.8   & 582557709 & 582706836.6 & 7.8   & 582557709 & 582785381.2 & 8.3 \\
     
    10C1k.6 & 581100882.1 & 582046361.4 & 5.3   & 580986510.1 & 581062186.1 & 7.5   & 580986510.1 & 581113261.8 & 7.6 \\
     
    10C1k.7 & 343334932.8 & 343819864.2 & 5.4   & 343312412.6 & 343395225.6 & 6.8   & 343312412.6 & 343503363 & 7.9 \\
     
    10C1k.8 & 564506471.2 & 564723479.2 & 5.1   & 564496866.9 & 564663214.5 & 7.5   & 564496866.9 & 564804840.7 & 7.5 \\
     
    10C1k.9 & 423567170.5 & 423891849.3 & 9.1   & 423562402.5 & 423582414.7 & 7.1   & 423562402.5 & 423618392.5 & 7.3 \\
    \bottomrule
    \end{tabular}%
  \label{tab:Results_Type2_3Algorithms}%
\end{table*}%

    \vspace{5mm}
\begin{table*}[htbp]
  \centering
  \caption{Results Obtained By GACSPT, N-LSEA and M-LSEA for Instances of Type 3}
    \begin{tabular}{l r r r r r r r r r}
    	\toprule
    	\multicolumn{1}{c}{\multirow{2}[4]{*}{\textbf{Instances}}} & 
    	\multicolumn{3}{c}{\textbf{GACSPT}} &
    	\multicolumn{3}{c}{\textbf{M-LSEA}} & 
    	\multicolumn{3}{c}{\textbf{N-LSEA}} \\
    	
    	\cmidrule(l{3pt}r{3pt}){2-4} \cmidrule(l{3pt}r{3pt}){5-7}  \cmidrule(l{3pt}r{3pt}){8-10} 
    	& 
    	\multicolumn{1}{c}{\textbf{BF}} & 
    	\multicolumn{1}{c}{\textbf{Avg}} & 
    	\multicolumn{1}{c}{\textbf{Rm}} & 
    	
    	\multicolumn{1}{c}{\textbf{BF}} & 
    	\multicolumn{1}{c}{\textbf{Avg}} & 
    	\multicolumn{1}{c}{\textbf{Rm}} & 
    	
    	\multicolumn{1}{c}{\textbf{BF}} & 
    	\multicolumn{1}{c}{\textbf{Avg}} & 
    	\multicolumn{1}{c}{\textbf{Rm}} \\
    	
    	\midrule
    	
	    6i300 & 19264.5 & 19278.3 & 0.9   & 19264.5 & 19265.0 & 0.3   & 19264.5 & 19264.5 & 0.3 \\
	    
	    6i350 & 21217.2 & 21236.3 & 1.1   & 21217.2 & 21217.2 & 0.4   & 21217.2 & 21217.2 & 0.4 \\
	    
	    6i400 & 29348.2 & 29350.1 & 1.0   & 29348.2 & 29348.2 & 0.5   & 29348.2 & 29348.4 & 0.5 \\
	    
	    6i450 & 35681.5 & 35684.2 & 1.3   & 35681.5 & 35682.5 & 0.7   & 35681.5 & 35682.8 & 0.7 \\
	    
	    6i500 & 37510.1 & 37516.0 & 2.1   & 37510.1 & 37510.7 & 0.8   & 37510.1 & 37510.6 & 0.8 \\
	    \bottomrule
    \end{tabular}%
  \label{tab:Results_Type3_3Algorithms}%
\end{table*}%

\begin{table*}[htbp]
  \centering
  \caption{Results Obtained By GACSPT, N-LSEA and M-LSEA for Instances of Type 4}
   \begin{tabular}{l r r r r r r r r r}
   	\toprule
   	\multicolumn{1}{c}{\multirow{2}[4]{*}{\textbf{Instances}}} & 
   	\multicolumn{3}{c}{\textbf{GACSPT}} &
   	\multicolumn{3}{c}{\textbf{M-LSEA}} & 
   	\multicolumn{3}{c}{\textbf{N-LSEA}} \\
   	
   	\cmidrule(l{3pt}r{3pt}){2-4} \cmidrule(l{3pt}r{3pt}){5-7}  \cmidrule(l{3pt}r{3pt}){8-10} 
   	& 
   	\multicolumn{1}{c}{\textbf{BF}} & 
   	\multicolumn{1}{c}{\textbf{Avg}} & 
   	\multicolumn{1}{c}{\textbf{Rm}} & 
   	
   	\multicolumn{1}{c}{\textbf{BF}} & 
   	\multicolumn{1}{c}{\textbf{Avg}} & 
   	\multicolumn{1}{c}{\textbf{Rm}} & 
   	
   	\multicolumn{1}{c}{\textbf{BF}} & 
   	\multicolumn{1}{c}{\textbf{Avg}} & 
   	\multicolumn{1}{c}{\textbf{Rm}} \\
   	
   	\midrule
   	    
    4i200a & 97959.6 & 98392.2 & 0.4   & 97959.6 & 97959.6 & 0.1   & 97959.6 & 97959.6 & 0.1 \\
    
    4i200h & 87675.3 & 87778.9 & 0.4   & 87675.3 & 87675.3 & 0.1   & 87675.3 & 87675.3 & 0.1 \\
    
    4i200x1 & 123669.7 & 123751.3 & 0.4   & 123669.7 & 123669.7 & 0.1   & 123669.7 & 123669.7 & 0.1 \\
    
    4i200x2 & 114012.3 & 114060.3 & 0.4   & 114012.3 & 114012.3 & 0.1   & 114012.3 & 114012.3 & 0.1 \\
    
    4i200z & 131683.5 & 131703.4 & 0.4   & 131683.5 & 131683.5 & 0.1   & 131683.5 & 131683.5 & 0.1 \\
    \addlinespace
    4i400a & 214115.3 & 219283.5 & 1.5   & 214115.3 & 214115.3 & 0.3   & 214115.3 & 214115.3 & 0.3 \\
    
    4i400h & 256200.5 & 256272.7 & 1.4   & 256200.5 & 256200.5 & 0.3   & 256200.5 & 256200.5 & 0.3 \\
    
    4i400x1 & 188196.8 & 188408.0 & 1.5   & 188196.8 & 188196.8 & 0.3   & 188196.8 & 188196.8 & 0.3 \\
    
    4i400x2 & 159254.8 & 160086.6 & 1.8   & 159254.8 & 159254.8 & 0.3   & 159254.8 & 159254.8 & 0.3 \\
    
    4i400z & 221423.9 & 221592.7 & 1.1   & 221423.9 & 221423.9 & 0.3   & 221423.9 & 221423.9 & 0.3 \\
    \bottomrule

    \end{tabular}%
  \label{tab:Results_Type4_3Algorithms}%
\end{table*}%

\end{landscape}

When the number of clusters is small, the search space for upper level is not too large. Consequently, the local search algorithm for upper level can converge to the solution which is near the global optimum. On the other hand, when the number of clusters increases, the performance of GACSPT outperforms \gls{m-lsea}. As we can figure out from the Figure~\ref{fig:13}, the biggest gap recorded in this group is 12.6 belong to the instance 100rat783-10x10 and the smallest gap obtained is 0.02 belong the test 15i500-306. The instances 20i2500-706 and \text{200i2500-710} have the same number of vertices while the number of clusters is 10 times difference, but the value PI we obtained are 1.17 and 6.67. A conclusion could be drawn form this result is that the number of clusters directly affect to the performance of our algorithms. The graph in Figure~\ref{fig:12} clarifies this conclusion. In this graph, we choose some instances of which the numbers of clusters are between 5 and 25. When the number of clusters is less than 10, \gls{m-lsea} returns the better results in comparison with GACSPT. On the other hand, GACSPT outperforms \gls{m-lsea} when the number of clusters is greater or equal to 15.

\begin{minipage}[b][][b]{.46\linewidth}
	\renewcommand{\scalefigure}{0.68}
	\begin{figure}[H]
		\includegraphics[scale=\scalefigure]{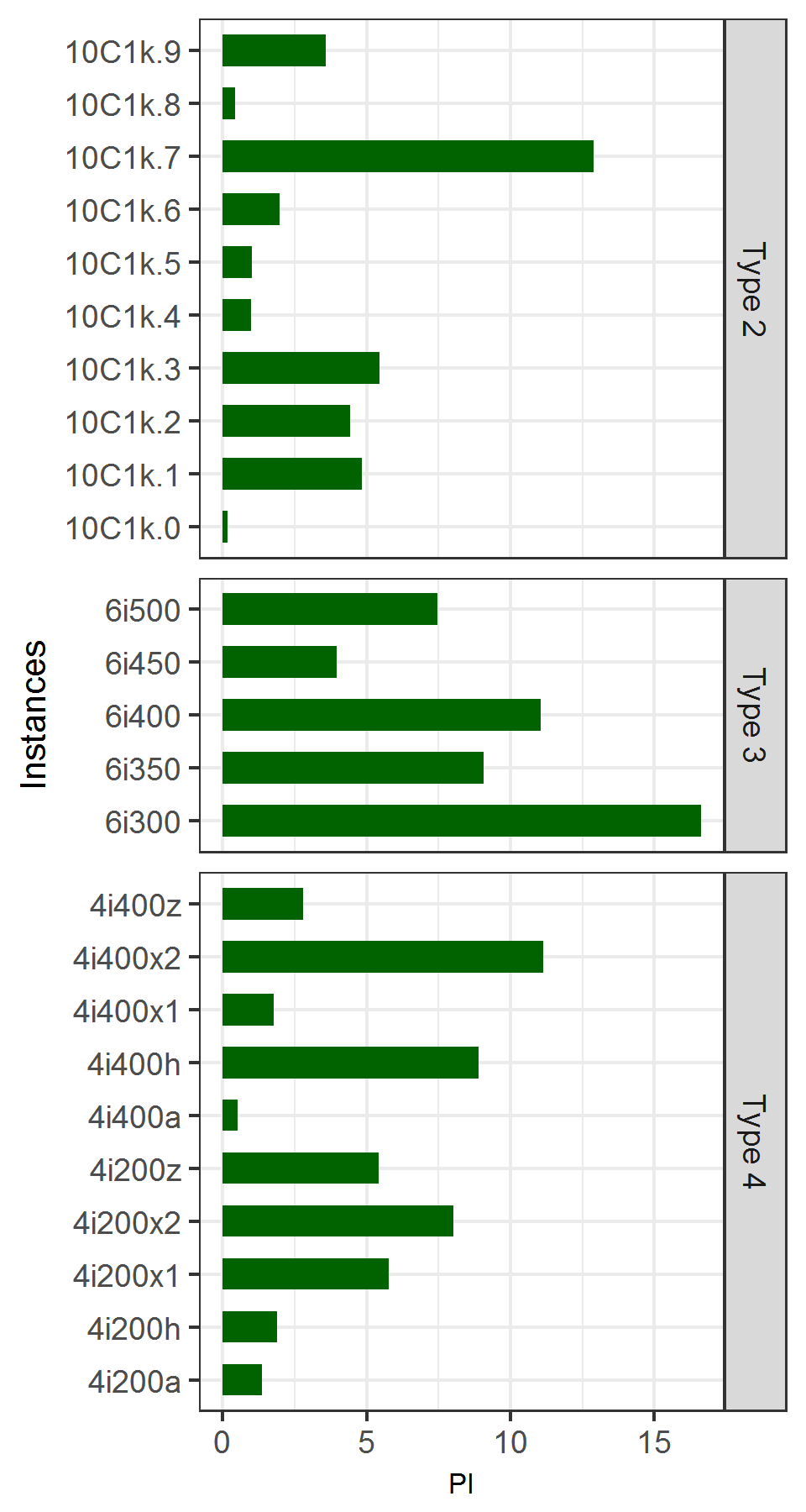}
		\caption{The performance of \gls{m-lsea} outperforms GACSPT on Euclidean instances in Type 2, Type 3 and Type 4.}
		\label{fig:11}
	\end{figure}
\end{minipage}
\hfill
\begin{minipage}[b][][b]{.46\linewidth}
	\renewcommand{\scalefigure}{0.68}
	\begin{figure}[H]
		\includegraphics[scale=\scalefigure]{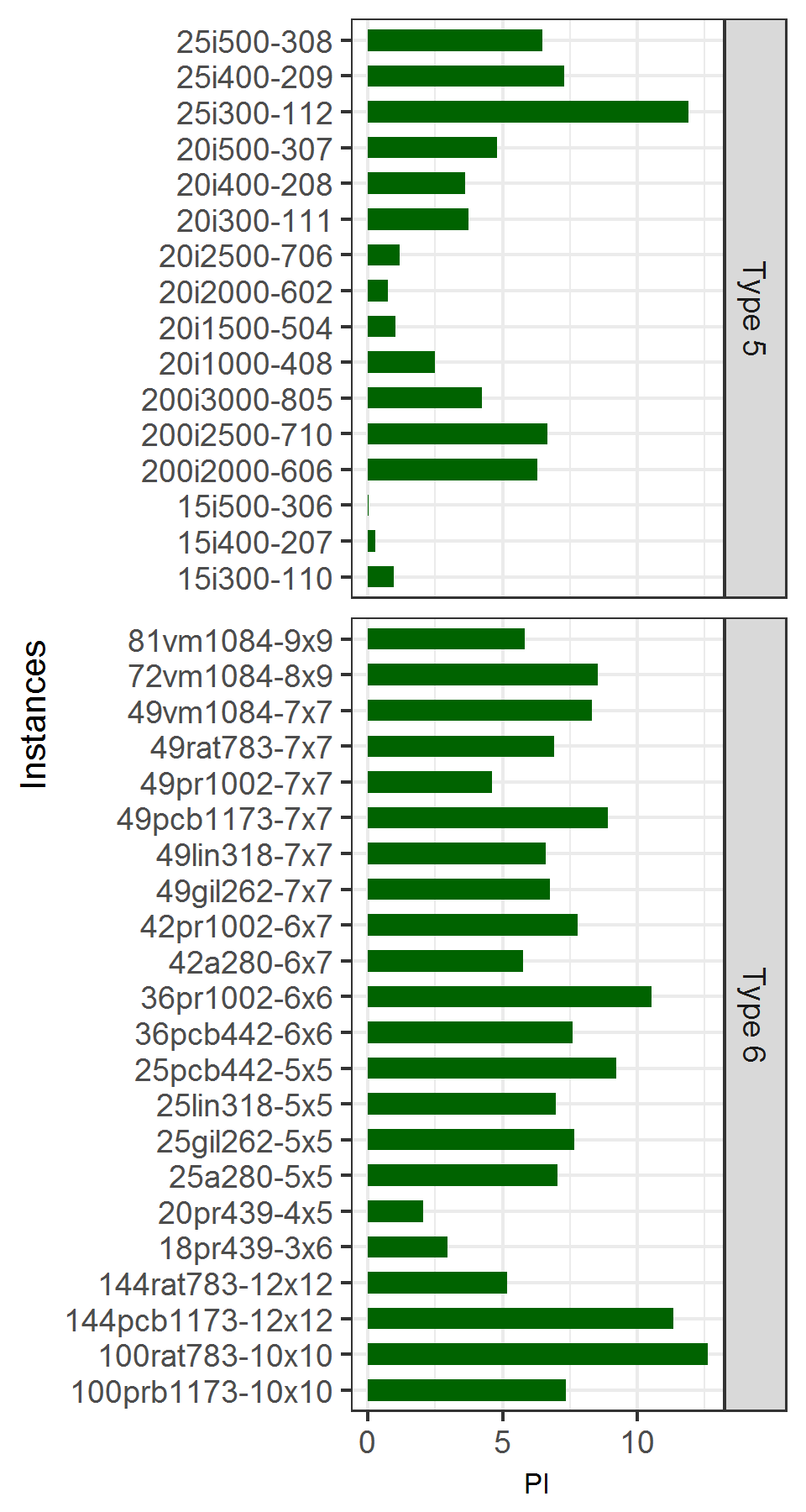}
		\caption{The performance of GACSPT outperforms \gls{m-lsea} for Euclidean instances of Type 5 and Type 6 }
		\label{fig:13}
	\end{figure}
\end{minipage}

\renewcommand{\scalefigure}{0.68}
\begin{figure}
	\centering
	\includegraphics[scale=\scalefigure]{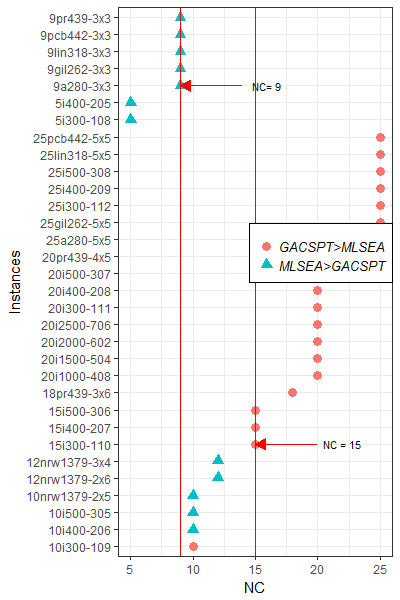}
	\caption{The relationship between number clusters and the performance of \gls{m-lsea} comparing to GACSPT (\gls{m-lsea} $>$ GACSPT mean that \gls{m-lsea} outperform GACSPT; GACSPT $>$ \gls{m-lsea} mean that GACSPT outperform \gls{m-lsea}; NC: Number of Cluster)}
	\label{fig:12}
\end{figure}

\subsubsection{Comparison the performances of \gls{m-lsea} and \gls{n-lsea}}

The running time of M-LSEA is equal to that of N-LSEA for almost all instances. In some cases,\gls{n-lsea} runs to the optimum faster than \gls{m-lsea}. The reason for this result is that when apply multitasking evolutionary algorithms for lower level tasks, we can find the solutions which are more optimal then single evolutionary algorithms. Thereafter, the local search for upper level in \gls{m-lsea} can go further than \gls{n-lsea}, and can go to the more optimal solution. The experimental results show that with regard to instances with small number of clusters, \gls{m-lsea} outperforms \gls{n-lsea} for almost all instances. As we can figure out in those tables below, the performance of \gls{m-lsea} is better than\gls{n-lsea} in almost all instances of Type 2 (in Table~\ref{tab:Results_Type2_3Algorithms}), Type 3 (in Table~\ref{tab:Results_Type3_3Algorithms}), which the number clusters are smaller than 11. In those tests, the biggest value PI obtained is 0.03 from the instance 10C1k.3. In some instances, \gls{n-lsea} defeat \gls{m-lsea}, but the PI value obtained is very trivial, nearly 0.002. Table~\ref{tab:Results_Type4_3Algorithms} presents results obtained by algorithms on instances of Type 4, those results show that when the number clusters of all instances are 4, the results we obtain are the same for both algorithm \gls{m-lsea} and \gls{n-lsea}. Furthermore, the best value and the average are equal for all tests in this type. Based on this result, we can see the good performance of the bi-level model for \gls{cluspt}.

\begin{landscape} 
\begin{table*}[htbp]
  \centering
  \caption{Results Obtained By GACSPT, N-LSEA and M-LSEA for Instances of Type 1 Large}
  \begin{tabular}{l r r r r r r r r r}
    	\toprule
    	\multicolumn{1}{c}{\multirow{2}[4]{*}{\textbf{Instances}}} & 
    	\multicolumn{3}{c}{\textbf{GACSPT}} &
    	\multicolumn{3}{c}{\textbf{M-LSEA}} & 
    	\multicolumn{3}{c}{\textbf{N-LSEA}} \\
    	
    	\cmidrule(l{3pt}r{3pt}){2-4} \cmidrule(l{3pt}r{3pt}){5-7}  \cmidrule(l{3pt}r{3pt}){8-10} 
    	& 
    	\multicolumn{1}{c}{\textbf{BF}} & 
    	\multicolumn{1}{c}{\textbf{Avg}} & 
    	\multicolumn{1}{c}{\textbf{Rm}} & 
    	
    	\multicolumn{1}{c}{\textbf{BF}} & 
    	\multicolumn{1}{c}{\textbf{Avg}} & 
    	\multicolumn{1}{c}{\textbf{Rm}} & 
    	
    	\multicolumn{1}{c}{\textbf{BF}} & 
    	\multicolumn{1}{c}{\textbf{Avg}} & 
    	\multicolumn{1}{c}{\textbf{Rm}} \\
    	
    	\midrule
	 	     
		 10a280 & 27925.2 & 27985.2 & 0.89  & 28045.1 & 28055.7 & 0.58  & 28045.1 & 28060.6 & 0.58 \\
		  
		 10gil262 & 27637.5 & 27671.7 & 0.77  & 27637.5 & 27660.7 & 0.50  & 27637.5 & 27669.7 & 0.57 \\
		  
		 10lin318 & 809750.0 & 809913.0 & 0.99  & 809750.0 & 810078.5 & 0.71  & 809750.0 & 810089.7 & 0.77 \\
		  
		 10pcb442 & 741195.9 & 742259.5 & 1.86  & 741195.8 & 741812.0 & 1.43  & 741195.8 & 742329.0 & 1.40 \\
		  
		 10pr439 & 1904690.2 & 1905689.2 & 1.70  & 1904690.2 & 1907047.0 & 1.47  & 1904690.2 & 1907405.1 & 1.52 \\
		  \addlinespace
		 150nrw1379 & 1684591.2 & 1906257.2 & 6.27  & 2008530.6 & 2112865.2 & 13.83 & 2242466.5 & 2334146.5 & 6.90 \\
		  
		 150pcb1173 & 2155219.0 & 2375069.6 & 5.69  & 2416570.6 & 2500536.2 & 8.75  & 2606021.9 & 2700678.7 & 6.35 \\
		  
		 150pr1002 & 8668720.0 & 9417483.2 & 5.27  & 9803738.5 & 10222390.4 & 7.34  & 10723866.2 & 11192116.1 & 5.46 \\
		  
		 150rat783 & 258228.8 & 275243.6 & 4.62  & 283213.2 & 298043.5 & 4.59  & 315040.0 & 324205.6 & 3.89 \\
		  
		 150vm1084 & 9950266.8 & 11028610.5 & 5.81  & 11845724.8 & 12349470.2 & 9.17  & 12676235.7 & 13540391.5 & 9.48 \\
		  \addlinespace
		 25a280 & 29924.4 & 30893.9 & 0.90  & 32415.0 & 33503.5 & 0.17  & 32223.4 & 33576.3 & 0.21 \\
		  
		 25gil262 & 30348.8 & 30922.9 & 0.77  & 31754.4 & 33119.9 & 0.20  & 32155.0 & 33281.0 & 0.19 \\
		  
		 25lin318 & 586178.0 & 606303.5 & 1.16  & 630959.1 & 648762.7 & 0.25  & 625209.4 & 650096.4 & 0.32 \\
		  
		 25pcb442 & 741254.6 & 763288.5 & 1.74  & 802807.3 & 842530.6 & 0.43  & 801798.2 & 848188.8 & 0.57 \\
		  
		 25pr439 & 1517343.9 & 1564461.0 & 1.81  & 1625683.4 & 1697265.7 & 0.30  & 1644541.0 & 1706810.8 & 0.42 \\
		  \addlinespace
		 50a280 & 37338.4 & 39522.3 & 1.09  & 40156.2 & 41872.7 & 0.20  & 41578.2 & 43692.4 & 0.25 \\
		  
		 50gil262 & 27076.5 & 29095.3 & 1.01  & 30015.2 & 31578.4 & 0.17  & 31050.2 & 33122.5 & 0.26 \\
		  
		 50lin318 & 702110.7 & 732497.7 & 1.27  & 755164.1 & 785033.9 & 0.26  & 774567.9 & 812480.7 & 0.35 \\
		  
		 50nrw1379 & 1886278.3 & 1950026.9 & 9.36  & 2004269.1 & 2061233.7 & 4.38  & 2050614.9 & 2123237.9 & 5.33 \\
		  
		 50pcb1173 & 1213167.4 & 1297153.5 & 6.61  & 1348291.8 & 1416301.0 & 2.50  & 1435151.7 & 1501454.6 & 3.96 \\
		  \addlinespace
		 50pcb442 & 936589.1 & 984005.5 & 1.47  & 1033040.9 & 1054291.4 & 0.35  & 1060758.0 & 1091788.4 & 0.64 \\
		  
		 50pr1002 & 5419480.2 & 5969365.8 & 3.38  & 6161282.7 & 6537955.6 & 1.81  & 6482992.0 & 6926391.9 & 2.52 \\
		  
		 50pr439 & 2209366.5 & 2328238.2 & 1.32  & 2379043.1 & 2465838.0 & 0.39  & 2493240.2 & 2567883.5 & 0.58 \\
     	\bottomrule
    \end{tabular}%
  \label{tab:Results_Type1Large_3Algorithms}%
\end{table*}%

\end{landscape}

In Type 1, Type 4, Type 5 of Euclidean instances, the number of cluster fluctuates from 5 to 200. In almost all instances of those types, the performance of \gls{m-lsea} is better than \gls{n-lsea}. Results obtained by 2 algorithms in Table~\ref{tab:Results_Type1Large_3Algorithms} show that \gls{m-lsea} outperforms \gls{m-lsea} on all instances in Type 1. The highest gap with which \gls{m-lsea} outperforms \gls{m-lsea} is 9.48 of the instance 150pcb1173, and the smallest belongs to the test 10lin318, 0.0013. We can figure out from the Figure~\ref{fig:performance-M-LSEA-N-LSEA-Euclid-Type1} that when the number of cluster increases, the PI value obtained is bigger. The experimental results in Table~\ref{tab:Results_Type5Large_3Algorithms} show that in Type 5, there is exact one problem instance 10i500-305 for which \gls{m-lsea} does not outperform \gls{m-lsea} (value in bold), and the value of PI(\gls{m-lsea}, \gls{n-lsea}) for it is very small, -0.0095. In Figure~\ref{fig:performance-M-LSEA-N-LSEA-Euclid-Type5-6}, the value PI of instances which has the number of clusters from 5 to 25 are small, however, when the number of clusters is 200, we can see this value PI increases suddenly. 

Similar to Type 1 and Type 5, the results in Table~\ref{tab:Results_Type6Large_3Algorithms} show that \gls{m-lsea} outperform \gls{n-lsea} for almost all instances of Type 6, only three problem instances (9a280-3x3, 9gil262-3x3 and 9pr439-3x3) are opposite (values in bold), but the value PI obtained of those tests are trivial (from -0.02 to -0.01). To see clearly the affection of number of clusters to the performance of multitasking evolutionary, we draw the Figure~\ref{fig:Relationship-Between-NumberCluster-PI-M-LSEA-N-LSEA} of the relation between the number of cluster and the averaged value PI for all instances with same number of cluster in Type 1. Based on this graph, we can clarify the conclusion mentioned above.

\begin{landscape}    
    \begin{table*}[htbp]
  \centering
  \caption{Results Obtained By GACSPT, \gls{n-lsea} and \gls{m-lsea} for Instances of Type 5 Large}
  \begin{tabular}{l r r r r r r r r r}
  	\toprule
  	\multicolumn{1}{c}{\multirow{2}[4]{*}{\textbf{Instances}}} & 
  	\multicolumn{3}{c}{\textbf{GACSPT}} &
  	\multicolumn{3}{c}{\textbf{M-LSEA}} & 
  	\multicolumn{3}{c}{\textbf{N-LSEA}} \\
  	
  	\cmidrule(l{3pt}r{3pt}){2-4} \cmidrule(l{3pt}r{3pt}){5-7}  \cmidrule(l{3pt}r{3pt}){8-10} 
  	& 
  	\multicolumn{1}{c}{\textbf{BF}} & 
  	\multicolumn{1}{c}{\textbf{Avg}} & 
  	\multicolumn{1}{c}{\textbf{Rm}} & 
  	
  	\multicolumn{1}{c}{\textbf{BF}} & 
  	\multicolumn{1}{c}{\textbf{Avg}} & 
  	\multicolumn{1}{c}{\textbf{Rm}} & 
  	
  	\multicolumn{1}{c}{\textbf{BF}} & 
  	\multicolumn{1}{c}{\textbf{Avg}} & 
  	\multicolumn{1}{c}{\textbf{Rm}} \\
  	
  	\midrule
    10i300-109 & 112681.01 & 112727.02 & 0.9   & 112681.0 & 112735.7 & 0.8   & 112681.0 & 112836.3 & 0.8 \\
     
    10i400-206 & 207521.68 & 207680.73 & 1.5   & 207521.7 & 207659.0 & 1.2   & 207521.7 & 207754.8 & 1.3 \\
     
    10i500-305 & 349676.63 & 349867.4 & 2.3   & 349675.2 & \textbf{349786.9} & 1.8   & 349675.2 & 349753.7 & 2.1 \\
     
    15i300-110 & 112096.66 & 112171.58 & 1.0   & 112393.0 & 113266.0 & 1.1   & 112577.2 & 113625.4 & 1.0 \\
     
    15i400-207 & 164117.83 & 164639.69 & 1.8   & 164271.7 & 165102.8 & 2.1   & 164526.9 & 165633.3 & 1.9 \\
       \addlinespace
    15i500-306 & 300742.11 & 301622.71 & 2.3   & 300893.2 & 301683.8 & 3.2   & 301089.6 & 302133.3 & 3.2 \\
     
    200i2000-606 & 1891129.19 & 2092509.04 & 10.2  & 2158246.4 & 2232903.7 & 31.9  & 2321654.1 & 2384238.9 & 34.4 \\
     
    200i2500-710 & 1994094.42 & 2173749.26 & 12.4  & 2246518.0 & 2329187.5 & 47.8  & 2416474.0 & 2531295.8 & 49.6 \\
     
    200i3000-805 & 2255766.59 & 2604482.11 & 15.0  & 2642053.6 & 2719577.3 & 72.4  & 2794935.2 & 2965489.7 & 65.0 \\
     
    20i1000-408 & 468210.16 & 470540.71 & 6.4   & 474197.2 & 482607.7 & 7.5   & 476798.0 & 487856.2 & 6.8 \\
       \addlinespace
    20i1500-504 & 918090.86 & 922161.3 & 17.4  & 924919.7 & 931793.9 & 26.3  & 927656.0 & 938703.4 & 20.7 \\
     
    20i2000-602 & 1008231.17 & 1015629.22 & 32.3  & 1015882.6 & 1023325.8 & 49.8  & 1015831.5 & 1031809.5 & 38.3 \\
     
    20i2500-706 & 1374110.55 & 1383759.83 & 44.0  & 1387986.6 & 1400171.3 & 65.5  & 1385462.6 & 1404259.3 & 69.4 \\
     
    20i300-111 & 156347.69 & 158147.92 & 0.8   & 159748.9 & 164305.3 & 0.5   & 159485.6 & 165744.2 & 0.6 \\
     
    20i400-208 & 224023.77 & 225715.65 & 1.6   & 227590.9 & 234155.8 & 0.8   & 229122.5 & 236582.7 & 1.1 \\
       \addlinespace
    20i500-307 & 200332.04 & 202616.53 & 2.2   & 206036.9 & 212845.2 & 1.3   & 208000.2 & 214113.6 & 1.8 \\
     
    25i300-112 & 116282.21 & 118925.94 & 1.0   & 128229.1 & 134989.5 & 0.2   & 128379.8 & 136666.0 & 0.2 \\
     
    25i400-209 & 230554.75 & 235954.14 & 1.6   & 246372.0 & 254536.5 & 0.2   & 247441.7 & 255233.3 & 0.4 \\
     
    25i500-308 & 299528.75 & 302548.82 & 2.1   & 315900.0 & 323559.8 & 0.6   & 312869.2 & 326271.7 & 0.8 \\
     
    5i300-108 & 177185.93 & 177221.26 & 0.9   & 177185.9 & 177185.9 & 0.1   & 177185.9 & 177188.7 & 0.2 \\
       \addlinespace
    5i400-205 & 209389.84 & 209440.29 & 1.6   & 209389.8 & 209389.8 & 0.2   & 209389.8 & 209389.8 & 0.4 \\
     
    5i500-304 & 182024.65 & 182045.89 & 2.5   & 182024.0 & 182024.0 & 0.4   & 182024.0 & 182024.0 & 0.6 \\
    \bottomrule
    \end{tabular}%
  \label{tab:Results_Type5Large_3Algorithms}%
\end{table*}%

\end{landscape}

\begin{landscape}    
	\begin{table}[htbp]
  \centering
  \caption{Results Obtained By GACSPT, \glsentrytext{m-lsea} and \glsentrytext{m-lsea} for Instances of Type 6 Large} 
  \label{tab:Results_Type6Large_3Algorithms}%
    \begin{tabular}{l r r r r r r r r r}
    	\toprule
    	\multicolumn{1}{c}{\multirow{2}[4]{*}{\textbf{Instances}}} & 
    	\multicolumn{3}{c}{\textbf{GACSPT}} &
    	\multicolumn{3}{c}{\textbf{M-LSEA}} & 
    	\multicolumn{3}{c}{\textbf{N-LSEA}} \\
    	
    	\cmidrule(l{3pt}r{3pt}){2-4} \cmidrule(l{3pt}r{3pt}){5-7}  \cmidrule(l{3pt}r{3pt}){8-10} 
    	& 
    	\multicolumn{1}{c}{\textbf{BF}} & 
    	\multicolumn{1}{c}{\textbf{Avg}} & 
    	\multicolumn{1}{c}{\textbf{Rm}} & 
    	
    	\multicolumn{1}{c}{\textbf{BF}} & 
    	\multicolumn{1}{c}{\textbf{Avg}} & 
    	\multicolumn{1}{c}{\textbf{Rm}} & 
    	
    	\multicolumn{1}{c}{\textbf{BF}} & 
    	\multicolumn{1}{c}{\textbf{Avg}} & 
    	\multicolumn{1}{c}{\textbf{Rm}} \\
    	
    	\midrule
     
    100prb1173-10x10 & 1930392.0 & 2071547.2 & 4.8   & 2151529.8 & 2235551.1 & 6.1   & 2251880.5 & 2386566.3 & 6.9 \\
     
    100rat783-10x10 & 174016.4 & 190040.4 & 2.9   & 207391.3 & 217450.7 & 3.1   & 221520.1 & 238693.4 & 3.6 \\
     
    10nrw1379-2x5 & 1318517.0 & 1323558.2 & 24.2  & 1317301.4 & 1317826.0 & 16.2  & 1317301.4 & 1317923.3 & 16.3 \\
     
    12nrw1379-2x6 & 1243089.7 & 1257144.5 & 31.3  & 1239386.0 & 1241102.7 & 18.6  & 1239386.0 & 1241903.9 & 19.0 \\
     
    12nrw1379-3x4 & 1486829.2 & 1490375.5 & 33.7  & 1486754.7 & 1487880.3 & 21.0  & 1486729.5 & 1488527.1 & 19.3 \\
     \addlinespace
    144pcb1173-12x12 & 1650023.7 & 1825097.3 & 6.5   & 2004711.3 & 2058787.4 & 8.9   & 2104321.2 & 2252633.3 & 10.0 \\
     
    144rat783-12x12 & 282192.0 & 299267.7 & 3.6   & 302651.9 & 315549.3 & 5.1   & 327006.7 & 342296.8 & 5.3 \\
     
    18pr439-3x6 & 1472083.9 & 1480688.3 & 2.8   & 1495669.1 & 1526029.9 & 0.9   & 1499814.2 & 1527240.3 & 1.1 \\
     
    20pr439-4x5 & 1978155.0 & 1993787.0 & 2.8   & 2004829.9 & 2035944.8 & 1.3   & 2010075.1 & 2044879.2 & 1.3 \\
     
    25a280-5x5 & 41713.0 & 42362.9 & 1.0   & 43327.7 & 45577.6 & 0.4   & 43998.7 & 45831.2 & 0.4 \\
     \addlinespace
    25gil262-5x5 & 30711.9 & 31178.8 & 0.9   & 32847.7 & 33760.2 & 0.2   & 32995.8 & 34108.0 & 0.2 \\
     
    25lin318-5x5 & 715116.5 & 726620.3 & 1.3   & 764621.0 & 781227.2 & 0.4   & 758213.7 & 784246.3 & 0.4 \\
     
    25pcb442-5x5 & 740883.3 & 760869.8 & 2.0   & 808312.9 & 838117.2 & 0.7   & 822598.6 & 844904.8 & 0.6 \\
     
    36pcb442-6x6 & 863522.7 & 900900.4 & 2.0   & 935979.2 & 974970.9 & 0.6   & 964538.7 & 994331.8 & 0.5 \\
     
    36pr1002-6x6 & 5847061.5 & 6120588.8 & 6.8   & 6593220.0 & 6841454.0 & 3.0   & 6770556.5 & 6981861.9 & 2.7 \\
     \addlinespace
    42a280-6x7 & 44514.5 & 46047.7 & 1.1   & 47212.2 & 48859.3 & 0.3   & 48525.9 & 49654.5 & 0.3 \\
     
    42pr1002-6x7 & 7189930.4 & 7479862.3 & 5.9   & 7801486.9 & 8111203.7 & 3.1   & 7998019.2 & 8295123.1 & 2.9 \\
     
    49gil262-7x7 & 33156.2 & 34426.2 & 1.1   & 35565.4 & 36922.3 & 0.3   & 36850.9 & 38392.9 & 0.3 \\
     
    49lin318-7x7 & 586512.4 & 625703.5 & 1.4   & 651961.0 & 670013.3 & 0.4   & 676284.1 & 706475.8 & 0.4 \\
     
    49pcb1173-7x7 & 1436187.5 & 1532484.5 & 7.6   & 1605570.7 & 1682327.7 & 4.1   & 1673411.3 & 1754706.1 & 3.7 \\
     \addlinespace
    49pr1002-7x7 & 7329741.0 & 7626061.1 & 5.1   & 7699692.7 & 7993580.3 & 3.0   & 8003380.6 & 8290583.7 & 2.9 \\
     
    49rat783-7x7 & 236409.1 & 245741.3 & 2.7   & 259615.9 & 263980.9 & 1.9   & 264175.1 & 271575.3 & 1.7 \\
     
    49vm1084-7x7 & 7712747.2 & 8184598.3 & 7.2   & 8583738.4 & 8926994.6 & 3.9   & 8857520.0 & 9294158.2 & 2.9 \\
     
    72vm1084-8x9 & 7900283.8 & 8559406.1 & 5.3   & 9072194.5 & 9359547.1 & 5.0   & 9644992.0 & 9961561.0 & 3.5 \\
     
    81vm1084-9x9 & 9010431.2 & 9762762.4 & 4.4   & 10027522.0 & 10365001.5 & 4.8   & 10559177.4 & 10932291.4 & 3.9 \\
     \addlinespace
    9a280-3x3 & 28947.5 & 29041.8 & 0.7   & 28947.5 & \textbf{28973.3} & 0.5   & 28947.5 & 28966.0 & 0.5 \\
     
    9gil262-3x3 & 20935.9 & 20955.8 & 0.5   & 20935.9 & \textbf{20956.2} & 0.5   & 20935.9 & 20952.4 & 0.4 \\
     
    9lin318-3x3 & 716850.2 & 717151.1 & 0.6   & 717041.0 & 717586.3 & 0.7   & 717041.0 & 717920.5 & 0.5 \\
     
    9pcb442-3x3 & 760238.3 & 763669.9 & 1.7   & 760238.3 & \textbf{760536.8} & 1.2   & 760238.3 & 760939.9 & 1.0 \\
     
    9pr439-3x3 & 1800753.9 & 1805313.1 & 1.3   & 1800753.9 & 1803300.6 & 1.2   & 1800753.9 & 1802816.7 & 1.0 \\
     \bottomrule
    \end{tabular}%
  \end{table}%

\end{landscape}

\begin{minipage}[b][][b]{.46\linewidth}
	\renewcommand{\scalefigure}{0.68}
	\begin{figure}[H]
		\includegraphics[scale=\scalefigure]{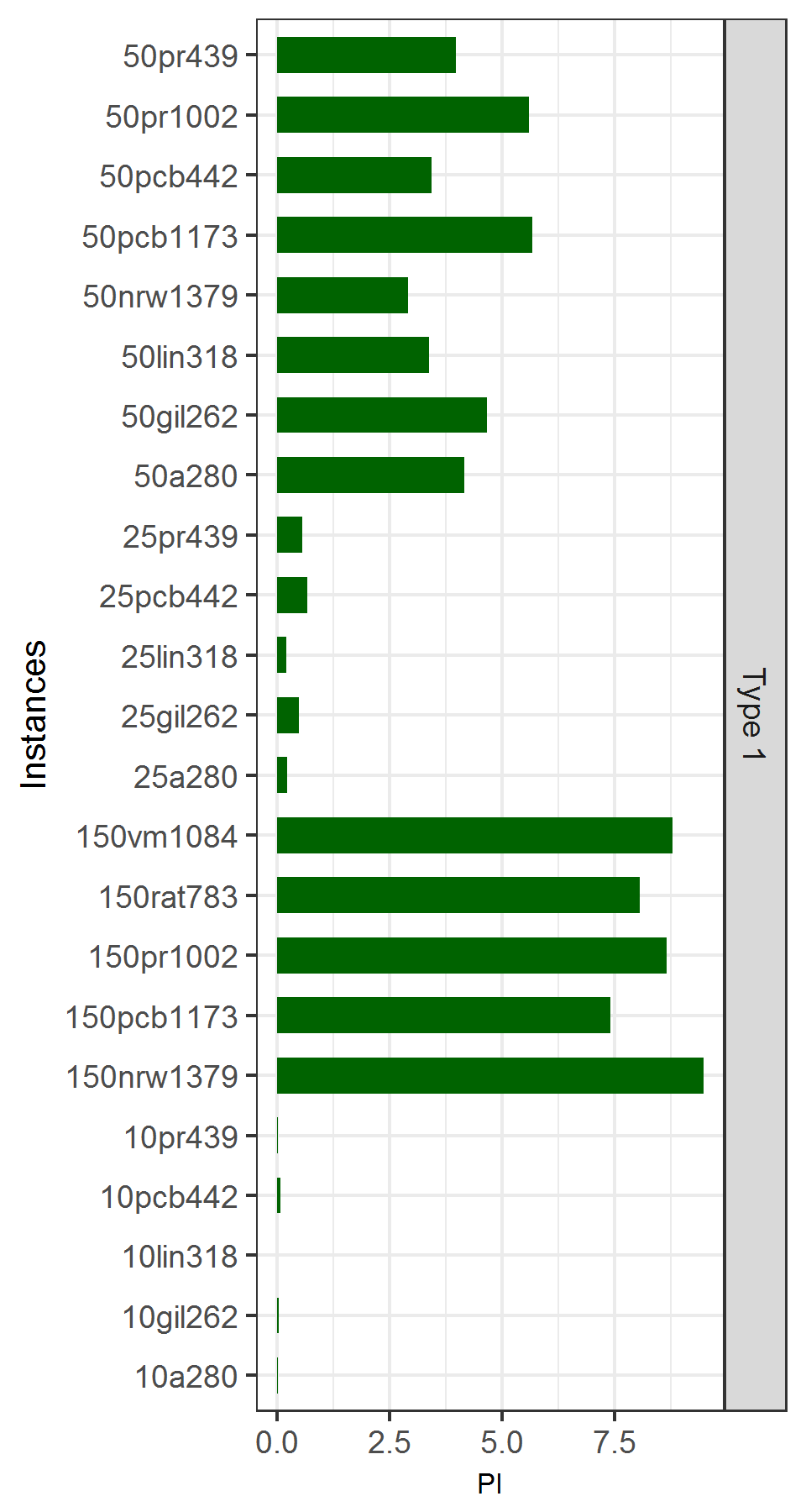}
		\caption{The performance of \gls{m-lsea} outperforms \gls{n-lsea} for Euclidean instances of Type 1.}
		\label{fig:performance-M-LSEA-N-LSEA-Euclid-Type1}
	\end{figure}
\end{minipage}
\hfill
\begin{minipage}[b][][b]{.46\linewidth}
	\renewcommand{\scalefigure}{0.68}
	\begin{figure}[H]
		\includegraphics[scale=\scalefigure]{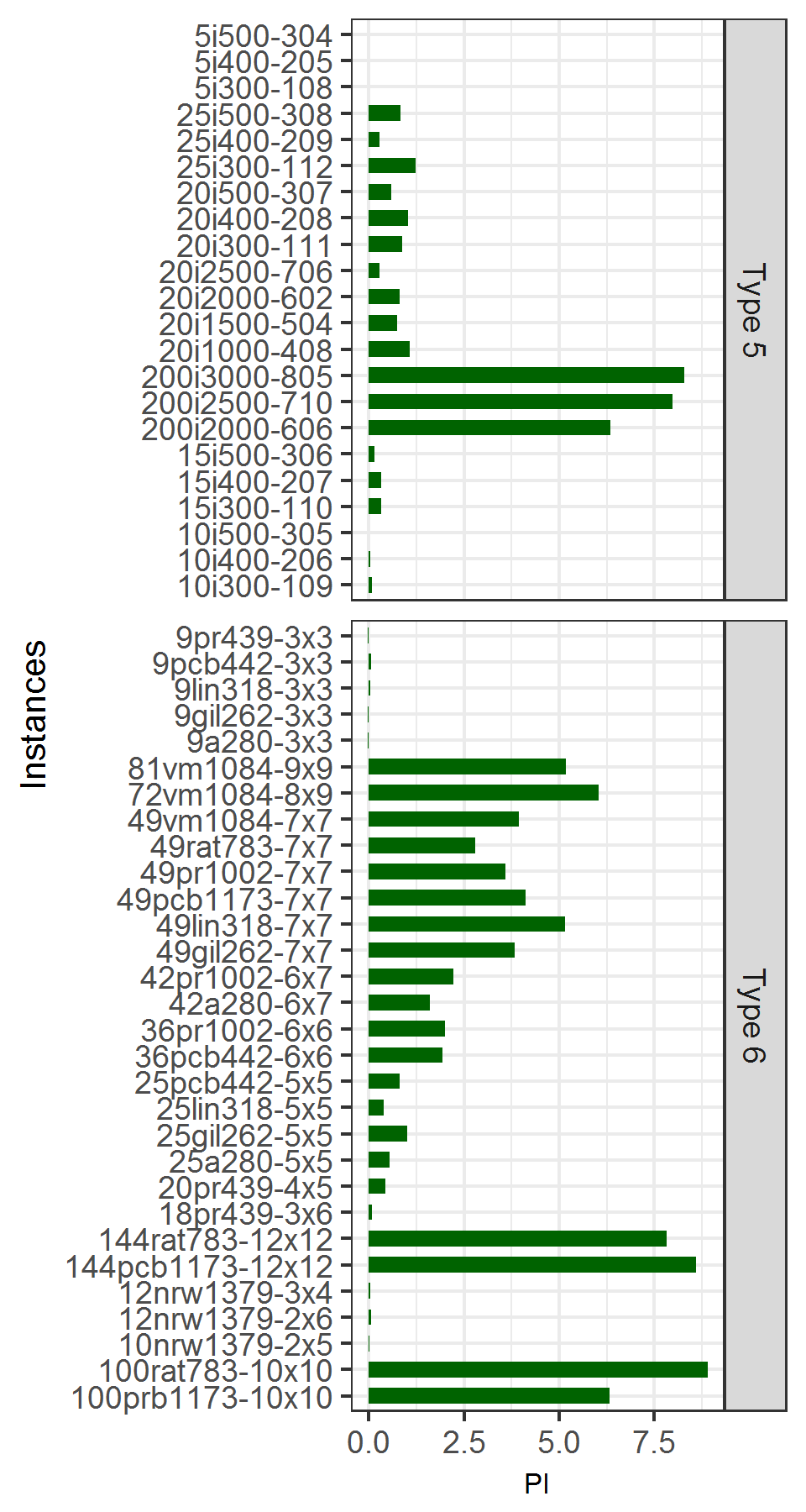}
		\caption{The performance of \gls{m-lsea} outperforms \gls{n-lsea} for Euclid Type 5 and Type 6 instances}
		\label{fig:performance-M-LSEA-N-LSEA-Euclid-Type5-6}
	\end{figure}
\end{minipage}

\setlength{\intextsep}{6pt}
\begin{figure}[htb]
   	\centering
    \includegraphics[width=0.45\textwidth]{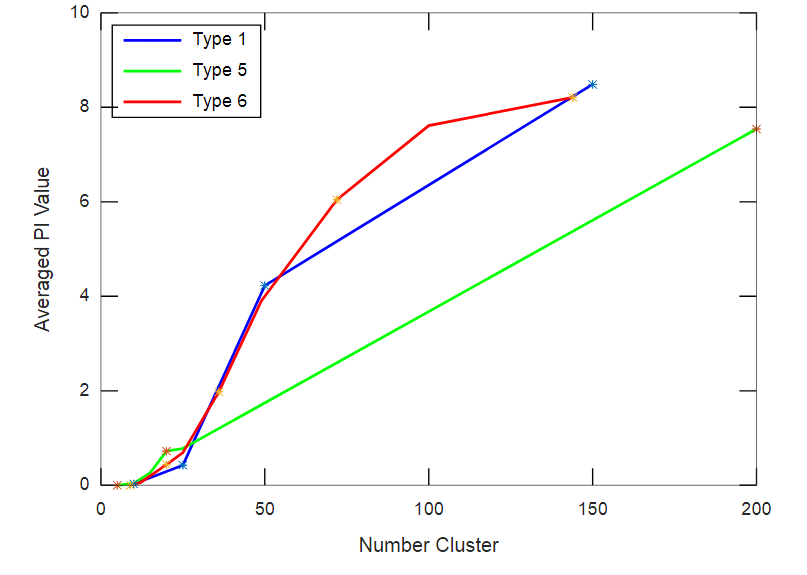}
    \caption{The relationship between the number of clusters and value PI between \gls{m-lsea} and \gls{n-lsea}}
    \label{fig:Relationship-Between-NumberCluster-PI-M-LSEA-N-LSEA}
\end{figure}

To prove the good performance of multitasking evolutionary, we simulate the convergence trends of \gls{m-lsea} and \gls{n-lsea} through graphs on Figure~\ref{fig:convergence-M-LSEA-N-LSEA-instances-10pcb442-25pcb442}, Figure~\ref{fig:convergence-M-LSEA-N-LSEA-instances-50pcb1173-150nrw1379}. Assume that we apply evolutionary algorithm for a lower task with an evolved population through generations, in the convergence trends presented hereafter, the objective function value of the fittest individual for task $T$ in generation $i$ is normalized as $\tilde{f_i} = (f_i - f_n) / (f_1 - f_n)$, which $f_i,f_n,f_1$ are the objective function values of the fittest individuals in generation $i, 1, n$,respectively. When apply bi-level optimization for \gls{cluspt}, we have to run evolutionary algorithm many times for all lower tasks, consequently, to exactly evaluate convergence trends, we average all normalized objective values in each generation.
The Figure~\ref{fig:convergence-M-LSEA-N-LSEA-instances-10pcb442-25pcb442}, Figure~\ref{fig:convergence-M-LSEA-N-LSEA-instances-50pcb1173-150nrw1379} depict the averaged convergence trends during the execution of \gls{m-lsea} and \gls{n-lsea} for instances of Type 1. It is clear that in Figure~\ref{fig:convergence-M-LSEA-N-LSEA-instances-10pcb442-25pcb442}, the curve corresponding to \gls{m-lsea} simulation surpasses \gls{n-lsea} for both instances 10pcb442 and 25pcb442. As the same individual representation and genetic operators, the improvement can be attributed to the exploitation of multiple lower task landscapes via genetic transfer, as is afforded by the evolutionary multitasking. In \gls{m-lsea}, the evolutionary multitasking is applied for two lower tasks whose upper candidates are neighbors, or almost similar, consequently, genetic materials are efficiently exchanged between two populations. Two instances 10pcb442 and 25pcb442 have the same number of vertices, but the number clusters are 10 and 25. We can figure out that algorithms for 10pcb442 converge faster than 25pcb442, the reason of this result is the search spaces for 10pcb442 when apply bi-level optimization are smaller than 25pcb442. The Figure~\ref{fig:convergence-M-LSEA-N-LSEA-instances-10pcb442-25pcb442} again demonstrates the effectiveness of \gls{m-lsea} in comparison to \gls{n-lsea}, in other words, evolutionary multitasking outperforms original evolutionary algorithm. The averaged convergence trends for Type 5 and Type 6 are given in Figure~\ref{fig:convergence-M-LSEA-N-LSEA-instances-50pcb1173-150nrw1379}. The efficacy of \gls{m-lsea} is again highlighted by the fast convergence trends in comparison to \gls{n-lsea}.

\renewcommand{\scalefigure}{0.3}
\begin{minipage}[b][][b]{.46\linewidth}
	\begin{figure}[H]
		\includegraphics[scale=\scalefigure]{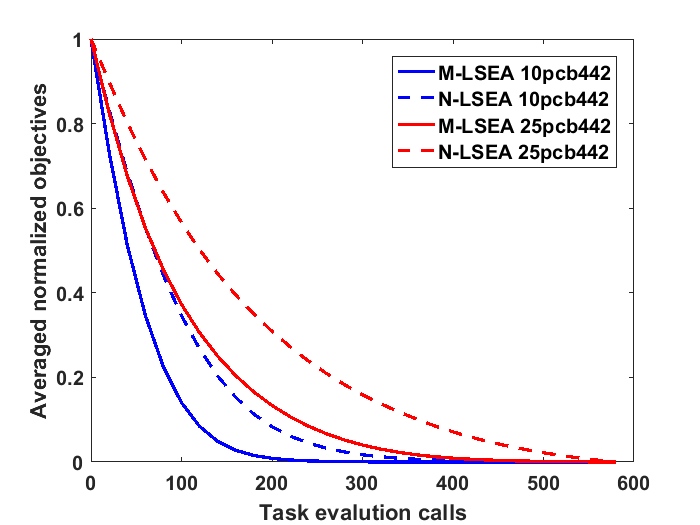}
		\caption{The convergence of \gls{m-lsea} and \gls{n-lsea} for instances 10pcb442 and 25pcb442}
		\label{fig:convergence-M-LSEA-N-LSEA-instances-10pcb442-25pcb442}
	\end{figure}
\end{minipage}
\hfill
\begin{minipage}[b][][b]{.46\linewidth}
	\begin{figure}[H]
		\includegraphics[scale=\scalefigure]{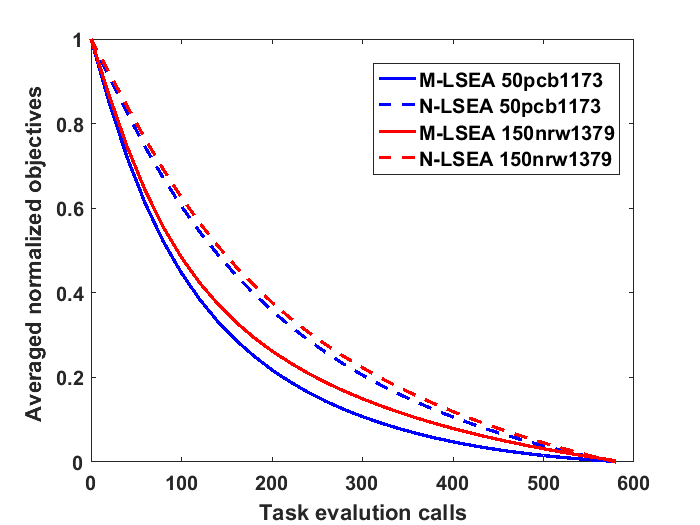}
		\caption{The convergence of \gls{m-lsea} and \gls{n-lsea} for instances 50pcb1173 and 150nrw1379}
		\label{fig:convergence-M-LSEA-N-LSEA-instances-50pcb1173-150nrw1379}
	\end{figure}
\end{minipage}

\subsubsection{Compare the performances of GACSP, \gls{n-lsea}, \gls{m-lsea} and EAM}
In this subsection, we analyze the \gls{rpd} between the average results obtained by the proposed heuristic algorithms and results obtained by exact algorithm on metric graph. 

A highlight in comparing the results received by algorithms GACSP, \gls{n-lsea}, \gls{m-lsea} with algorithm EAM is the ones which are obtained by two algorithms N-EA and M-EA on data sets of Types 2, 3, 4. On all instances in Type 3 and Type 4, the RPD value peaked at 0.00\% which means that solutions obtained by both algorithms \gls{n-lsea} and \gls{m-lsea} are very close to the  globally optimal solutions. On instances in Type 2, average RPD of NLSEA and \gls{m-lsea} were 0.04\% and 0.02\% respectively. The biggest RPD recorded in this type were only 0.06\% (for NLSEA) and 0.03\% (for \gls{m-lsea}). It means that these algorithms can produce solutions which are near  the optimal ones . More details of these algorithms in Type 3, Type 4 and Type 2 are presented in Table~\ref{tab:Results_Type3_RPD}, Table~\ref{tab:Results_Type4_RPD} and Table~\ref{tab:Results_Type2_RPD} respectively.

\begin{table}[htbp]
  \centering
  \caption{Relative Percentage Differences Between Results obtained by AEM and GACSPT, \gls{m-lsea}, \gls{m-lsea} for Instances of Type 3.}

    \begin{tabular}{l r r r r}
     \toprule
    \multicolumn{1}{c}{\textbf{Instances}} & 
    \multicolumn{1}{c}{\textbf{EAM}} & 
    \multicolumn{3}{c}{\textbf{RPD(.)}} \\
    
	\cmidrule(l{3pt}r{3pt}){3-5}
	&       & 
	\multicolumn{1}{c}{\textbf{GACSPT}} & 
	\multicolumn{1}{c}{\textbf{\gls{m-lsea}}} & 
	\multicolumn{1}{c}{\textbf{\gls{n-lsea}}} \\
    \midrule
    6i300 & 19264.5 & 0.07\% & 0.00\% & 0.00\% \\
     
    6i350 & 21217.2 & 0.09\% & 0.00\% & 0.00\% \\
     
    6i400 & 29348.2 & 0.01\% & 0.00\% & 0.00\% \\
     
    6i450 & 35681.5 & 0.01\% & 0.00\% & 0.00\% \\
     
    6i500 & 37510.1 & 0.02\% & 0.00\% & 0.00\% \\
     \bottomrule
    \end{tabular}%
  \label{tab:Results_Type3_RPD}%
\end{table}%

\begin{table}[htbp]
  \centering
  \caption{Relative Percentage Differences Between Results Obtained by AEM and GACSPT, \gls{n-lsea}, \gls{m-lsea} for Instances of Type 4.}
   \begin{tabular}{l r r r r}
   	\toprule
   	\multicolumn{1}{c}{\textbf{Instances}} & 
   	\multicolumn{1}{c}{\textbf{EAM}} & 
   	\multicolumn{3}{c}{\textbf{RPD(.)}} \\
   	
   	\cmidrule(l{3pt}r{3pt}){3-5}
   	&       & 
   	\multicolumn{1}{c}{\textbf{GACSPT}} & 
   	\multicolumn{1}{c}{\textbf{\gls{m-lsea}}} & 
   	\multicolumn{1}{c}{\textbf{\gls{n-lsea}}} \\
   	\midrule
   	
    4i200a & 97959.6 & 0.44\% & 0.00\% & 0.00\% \\
     
    4i200h & 87675.3 & 0.12\% & 0.00\% & 0.00\% \\
     
    4i200x1 & 123669.7 & 0.07\% & 0.00\% & 0.00\% \\
     
    4i200x2 & 114012.3 & 0.04\% & 0.00\% & 0.00\% \\
     
    4i200z & 131683.5 & 0.02\% & 0.00\% & 0.00\% \\
     \addlinespace
    4i400a & 214115.3 & 2.41\% & 0.00\% & 0.00\% \\
     
    4i400h & 256200.5 & 0.03\% & 0.00\% & 0.00\% \\
     
    4i400x1 & 188196.7 & 0.11\% & 0.00\% & 0.00\% \\
     
    4i400x2 & 159254.8 & 0.52\% & 0.00\% & 0.00\% \\
     
    4i400z & 221423.9 & 0.08\% & 0.00\% & 0.00\% \\
     \bottomrule

    \end{tabular}%
  \label{tab:Results_Type4_RPD}%
\end{table}%

\begin{table}[htbp]
  \centering
  \caption{Relative Percentage Differences Between Results obtained by AEM and GACSPT, \gls{n-lsea}, \gls{m-lsea} for Instances of Type 2.}
  \begin{tabular}{l r r r r}
  	\toprule
  	\multicolumn{1}{c}{\textbf{Instances}} & 
  	\multicolumn{1}{c}{\textbf{EAM}} & 
  	\multicolumn{3}{c}{\textbf{RPD(.)}} \\
  	
  	\cmidrule(l{3pt}r{3pt}){3-5}
  	&       & 
  	\multicolumn{1}{c}{\textbf{GACSPT}} & 
  	\multicolumn{1}{c}{\textbf{\gls{m-lsea}}} & 
  	\multicolumn{1}{c}{\textbf{\gls{n-lsea}}} \\
  	\midrule

    10C1k.0 & 603896769.1 & 0.13\% & 0.03\% & 0.05\% \\
     
    10C1k.1 & 555519829.5 & 0.13\% & 0.01\% & 0.02\% \\
     
    10C1k.2 & 740549053.6 & 0.06\% & 0.02\% & 0.04\% \\
     
    10C1k.3 & 594412757.2 & 0.03\% & 0.01\% & 0.04\% \\
     
    10C1k.4 & 533153746.9 & 0.10\% & 0.01\% & 0.02\% \\
     
    10C1k.5 & 582557709 & 0.05\% & 0.03\% & 0.04\% \\
     
    10C1k.6 & 580986510.1 & 0.18\% & 0.01\% & 0.02\% \\
     
    10C1k.7 & 343312412.6 & 0.15\% & 0.02\% & 0.06\% \\
     
    10C1k.8 & 564496866.9 & 0.04\% & 0.03\% & 0.05\% \\
     
    10C1k.9 & 423562402.5 & 0.08\% & 0.00\% & 0.01\% \\
     \bottomrule

    \end{tabular}%
  \label{tab:Results_Type2_RPD}%
\end{table}%

Table~\ref{tab:Results_Type5Small_RPD} shows the results obtained by the algorithms for instances of Type 5 Small. The results in Type 5 Small were slightly similar to those  in Type 2  in terms of the small RPD value of algorithms GACSP, \gls{n-lsea}, \gls{m-lsea}  with maximum RPD values of only 0.23\% (for GACSPT), 0.42\% (for \gls{n-lsea}) and 0.44\% (for \gls{m-lsea}). However, a main difference in comparing results of Type 5 Small with results of Type 2 is that of Type 5 Small, the results obtained by GACSP were slightly better than the ones obtained by both \gls{n-lsea} and \gls{m-lsea}. 

\begin{table}[htbp]
  \centering
    \caption{Relative Percentage Differences Between Results Obtained by EAM and GACSPT, \gls{n-lsea}, \gls{m-lsea} for Instances of Type 5 Small.}
   \begin{tabular}{l r r r r}
   	\toprule
   	\multicolumn{1}{c}{\textbf{Instances}} & 
   	\multicolumn{1}{c}{\textbf{EAM}} & 
   	\multicolumn{3}{c}{\textbf{RPD(.)}} \\
   	
   	\cmidrule(l{3pt}r{3pt}){3-5}
   	&       & 
   	\multicolumn{1}{c}{\textbf{GACSPT}} & 
   	\multicolumn{1}{c}{\textbf{\gls{m-lsea}}} & 
   	\multicolumn{1}{c}{\textbf{\gls{n-lsea}}} \\
   	\midrule
   	
    10i120-46 & 93925.0 & 0.02\% & 0.11\% & 0.12\% \\
     
    10i30-17 & 13276.6 & 0.04\% & 0.21\% & 0.20\% \\
     
    10i45-18 & 22890.4 & 0.01\% & 0.44\% & 0.38\% \\
     
    10i60-21 & 33694.8 & 0.11\% & 0.29\% & 0.26\% \\
     
    10i65-21 & 37353.1 & 0.23\% & 0.32\% & 0.42\% \\
     	 \addlinespace
     
    10i70-21 & 38059.5 & 0.07\% & 0.22\% & 0.37\% \\
     
    10i75-22 & 65361.9 & 0.03\% & 0.10\% & 0.20\% \\
     
    10i90-33 & 51931.2 & 0.02\% & 0.09\% & 0.16\% \\
     
    5i120-46 & 61451.5 & 0.01\% & 0.00\% & 0.00\% \\
     	 \addlinespace
     
    5i30-17 & 14399.9 & 0.00\% & 0.01\% & 0.00\% \\
     
    5i45-18 & 14884.3 & 0.00\% & 0.00\% & 0.00\% \\
     
    5i60-21 & 28422.7 & 0.00\% & 0.00\% & 0.04\% \\
     
    5i65-21 & 30907.8 & 0.00\% & 0.00\% & 0.00\% \\
     
    5i70-21 & 35052.8 & 0.00\% & 0.00\% & 0.00\% \\
     	 \addlinespace
     
    5i75-22 & 34692.5 & 0.01\% & 0.00\% & 0.01\% \\
     
    5i90-33 & 51977.0 & 0.00\% & 0.00\% & 0.00\% \\
     
    7i30-17 & 20438.9 & 0.00\% & 0.07\% & 0.11\% \\
     
    7i45-18 & 20512.0 & 0.01\% & 0.00\% & 0.00\% \\
     
    7i60-21 & 36263.9 & 0.01\% & 0.01\% & 0.08\% \\
     	 \addlinespace
     
    7i65-21 & 34847.6 & 0.00\% & 0.02\% & 0.13\% \\
     
    7i70-21 & 39487.6 & 0.01\% & 0.03\% & 0.04\% \\
    \bottomrule
    \end{tabular}%
  \label{tab:Results_Type5Small_RPD}%
\end{table}%

In other Types, although new algorithms produced good solutions with their RPD values of 0.00\% on some instances (5i300-108, 5i400-205, 5i500-304 of Type 5 Large; 5eil51, 5eil76, 5pr76, 5st70 of Type 1 Small; etc.), the RPD values of results obtained by these algorithms are large on some instances such as 85.0\% on instance 150nrw1379 for \gls{n-lsea}, 86.81\% on instance 200i2500-710 for \gls{n-lsea}. 

\setlength{\intextsep}{0pt}
\renewcommand{\scalefigure}{0.55}
\begin{figure}[htbp]
	\centering
	\begin{subfigure}[b]{.32\linewidth}
		\centering
		\includegraphics[scale=\scalefigure]{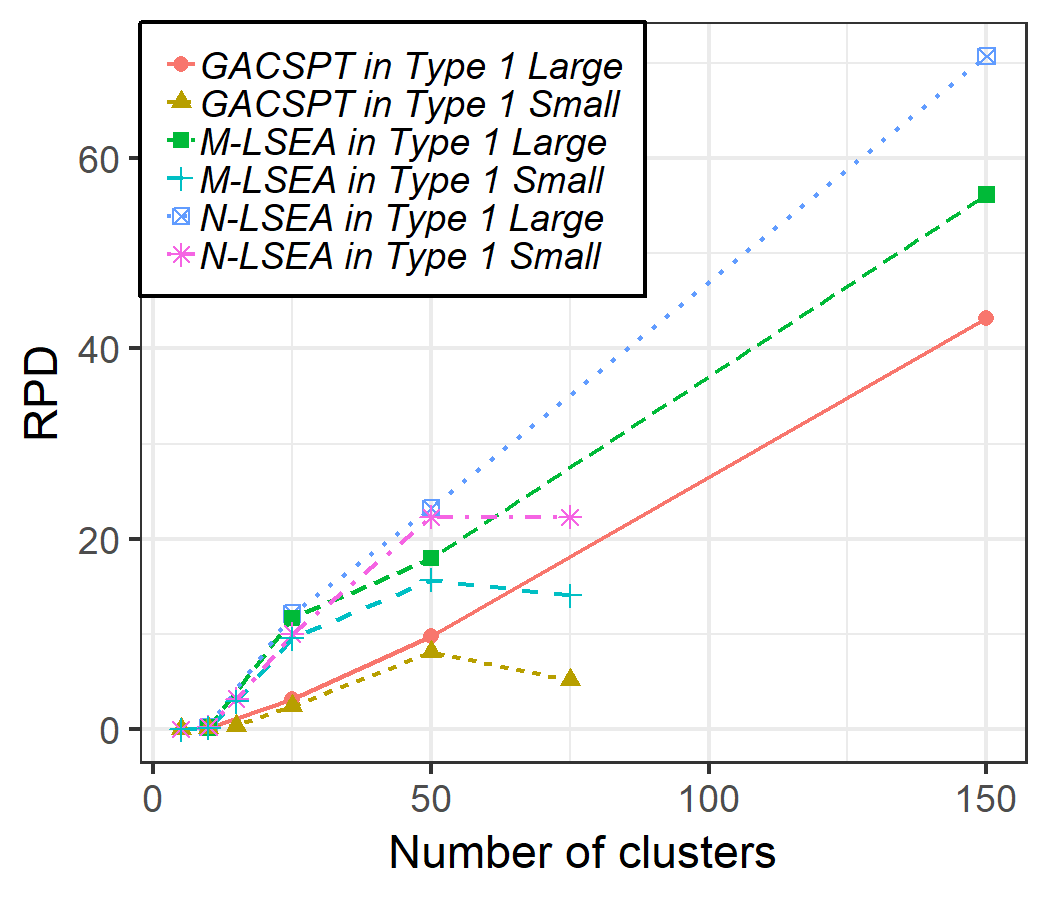}
		\caption{Type 1}
		\label{fig:RPD-Type1}
	\end{subfigure}
	\begin{subfigure}[b]{.32\linewidth}
		\centering
		\includegraphics[scale=\scalefigure]{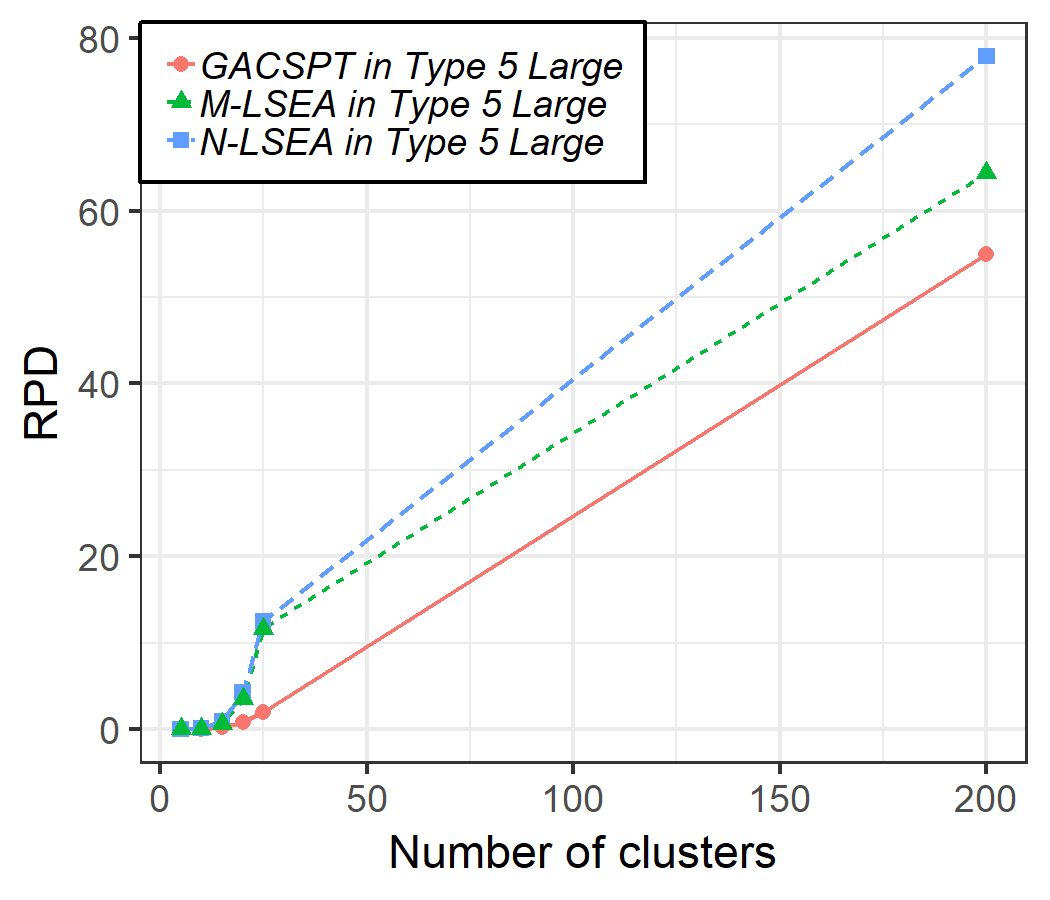}
		\caption{Type 5 Large}	
		\label{fig:RPD-Type5}
	\end{subfigure}
	\begin{subfigure}[b]{.32\linewidth}
		\centering
		\includegraphics[scale=\scalefigure]{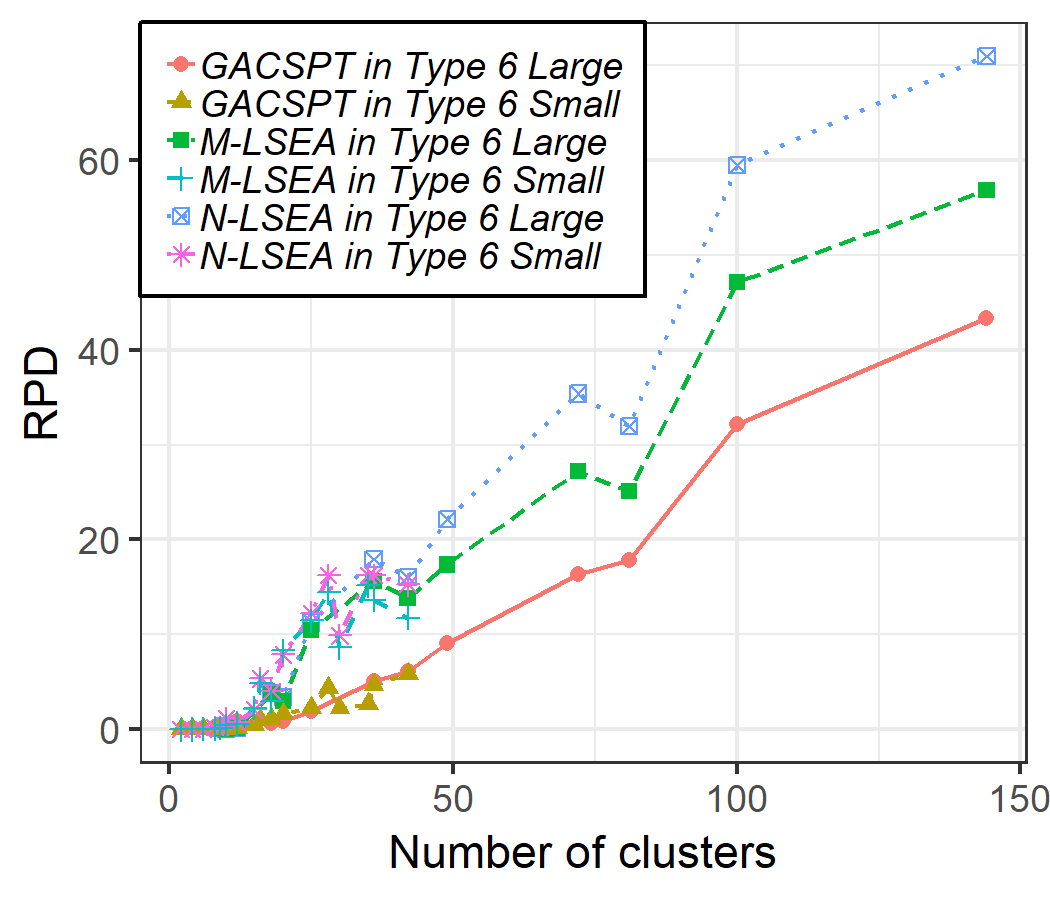}
		\caption{Type 6}
		\label{fig:RPD-Type6}
	\end{subfigure}
	\caption{The relationship between the number of clusters and RPD values of the new algorithms on instances in Type 1, Type~5~Large and Type 6}
	\label{fig:RPD-Type156-EUC}
\end{figure}
\setlength{\intextsep}{0pt}

Figure~\ref{fig:RPD-Type156-EUC} illustrates RPD values of the algorithms in Type 1, Type 5 and Type 6. From Figure~\ref{fig:RPD-Type156-EUC} it can be noted that the RPD value increases when the number of clusters on the instance increases. The reason behind this is that the new algorithms are applied to find solution of cluster graph H so when the dimensionality of the cluster graph H increases (in other word the dimensionality of the input problem increases), the quality of the solutions obtained by the new algorithms tend to decrease.

\section{Conclusion}
\label{Sec_Conclusion}
This paper proposed an exact algorithm for solving the \gls{cluspt} when the given input is a metric graph. For those arbitrary instances, \gls{cluspt} becomes an NP-hard problem. In order to enhance the performance of the search process when using approximate algorithms, this paper proposed two different ways to narrow down the search space by converting the original graph to much smaller one. On the first approach, the transformed graph is an undirected multi-graph whose edges set includes all edges between all pairs of clusters. Otherwise, the second approach changes the space of the previous approach to a bi-level space in which a candidate in the upper level offers a corresponding mutually exclusive search space for the lower level. Those proposed approaches boost the algorithm to be better in terms of running time and quality of solutions.

For solving the problem after transforming step of the first approach, we described an evolutionary algorithm to find solutions of \gls{cluspt} by focusing on optimizing a set of edges which connect among clusters. This technique only perform robustly on those test sets which have few edges between clusters. As a result, The second approach improves this disadvantage of the first approach by decomposing the space into many mutually exclusive search space and modeling the problem into a bi-level optimization problem. Accordingly, the \gls{n-lsea} is introduced to search for the optimal solution for this bi-level problem. The \gls{n-lsea} well-performs on those instances including small number clusters and many inter-cluster edges. From our observation for the lower level of\ gls{n-lsea}, the sub-problems in this level have high similarity which is indicated by the use of the local search in the upper level. This investigation convinces us to apply the idea of \gls{m-lsea} to take the advantage of meaningful building-blocks shared between different optimization tasks. The improvement in experimental results over \gls{n-lsea} via this multi-tasking scheme inspires the future works to apply gls{m-lsea} in graph-based problems, especially for those could be modeled into bi-level optimization.

Various types of problem datasets are selected to conduct experiments on the new algorithms. The experimental results demonstrated superior performance of the new algorithms in comparison with existing algorithms. An in-depth analysis also explained the impact of convergence trend and the number of clusters to the efficiency of the new algorithms. To enhance the performance of the novel algorithms, in the future, we will find a new mechanism to improve the quality of solutions on instances having large number of clusters.

\section*{Acknowledgements}
This research was sponsored by the U.S. Army Combat Capabilities Development Command (CCDC) Pacific and CCDC Army Research Laboratory (ARL) under Contract Number W90GQZ-93290007. The views and conclusions contained in this document are those of the authors and should not be interpreted as representing the official policies, either expressed or implied, of the CCDC Pacific and CCDC ARL and the U.S. Government. The U.S. Government is authorized to reproduce and distribute reprints for Government purposes notwithstanding any copyright notation hereon.

\bibliography{references}   

\end{document}